\documentclass[sigconf]{acmart}
\usepackage{xcolor}
\usepackage{caption}
\usepackage{subcaption}
\usepackage{algorithm,algcompatible,amsmath}
\usepackage[toc]{appendix}
\usepackage{comment}
\usepackage{multirow}
\usepackage{lineno}
\usepackage{float}

\AtBeginDocument{%
  \providecommand\BibTeX{{%
    \normalfont B\kern-0.5em{\scshape i\kern-0.25em b}\kern-0.8em\TeX}}}

\acmConference[KDD '24]{Knowledge Discovery and Data Mining}{August 25--29, 2024}{Barcelona, Spain}

\acmISBN{978-1-4503-XXXX-X/18/06}

\begin{document}
%\title{Lumos : On-Device Scene Text Recognition System That Enables Multimodal LLM To Read}
\title{\textsc{Lumos} : Empowering Multimodal LLMs with \\Scene Text Recognition}


\author{Ashish Shenoy}
\authornote{Joint First Authors}
\email{ashishvs@meta.com}
\affiliation{%
  \institution{Meta Reality Labs}
  \country{}
}

\author{Yichao Lu}
\authornotemark[1]
\email{yichaol@meta.com}
\affiliation{%
  \institution{Meta Reality Labs}
  \country{}
}

\author{Srihari Jayakumar}
\authornotemark[1]
\email{srihari2761@meta.com}
\affiliation{%
  \institution{Meta Reality Labs}
  \country{}
}

\author{Debojeet Chatterjee}
\authornotemark[1]
\email{debo@meta.com}
\affiliation{%
  \institution{Meta Reality Labs}
  \country{}
}

\author{Mohsen Moslehpour}
\authornotemark[1]
\email{moslehpour@meta.com}
\affiliation{%
  \institution{Meta Reality Labs}
  \country{}
}

\author{Pierce Chuang}
\authornotemark[1]
\email{pichuang@meta.com}
\affiliation{%
  \institution{Meta Reality Labs}
  \country{}
}

\author{Abhay Harpale}
\authornotemark[1]
\email{harpale@meta.com}
\affiliation{%
  \institution{Meta Reality Labs}
  \country{}
}

\author{Vikas Bhardwaj}
\authornotemark[1]
\email{vikasb@meta.com}
\affiliation{%
  \institution{Meta Reality Labs}
  \country{}
  }

\author{Di Xu}
\email{dixu@meta.com}
\affiliation{%
  \institution{Meta Reality Labs}
  \country{}
}

\author{Shicong Zhao}
\email{zsc@meta.com}
\affiliation{%
  \institution{Meta Reality Labs}
  \country{}
}

\author{Longfang Zhao}
\email{longfangzhao@meta.com}
\affiliation{%
  \institution{Meta Reality Labs}
  \country{}
}

\author{Ankit Ramchandani}
\email{ankit61@meta.com}
\affiliation{%
  \institution{Meta}
  \country{}}

\author{Xin Luna Dong}
\email{lunadong@meta.com}
\affiliation{%
  \institution{Meta Reality Labs}
  \country{}
}

\author{Anuj Kumar}
\email{anujk@meta.com}
\affiliation{%
  \institution{Meta Reality Labs}
  \country{}
}

\renewcommand{\shortauthors}{Shenoy, Lu, Jayakumar, Chatterjee, Moslehpour, Chuang, Harpale, Bhardwaj, et al.}

\begin{abstract}
  We introduce \textsc{Lumos}, the first end-to-end multimodal question-answering system with text understanding capabilities.  At the core of \textsc{Lumos} is a Scene Text Recognition (STR) component that extracts text from first person point-of-view images, the output of which is used to augment input to a Multimodal Large Language Model (MM-LLM). While building \textsc{Lumos}, we encountered numerous challenges related to STR quality, overall latency, and model inference. In this paper, we delve into those challenges, and discuss the system architecture, design choices, and modeling techniques employed to overcome these obstacles. We also provide a comprehensive evaluation for each component, showcasing high quality and efficiency.
\end{abstract}




\maketitle

\begin{figure*}[htp]
    \centering
    \includegraphics[width=0.8\textwidth]{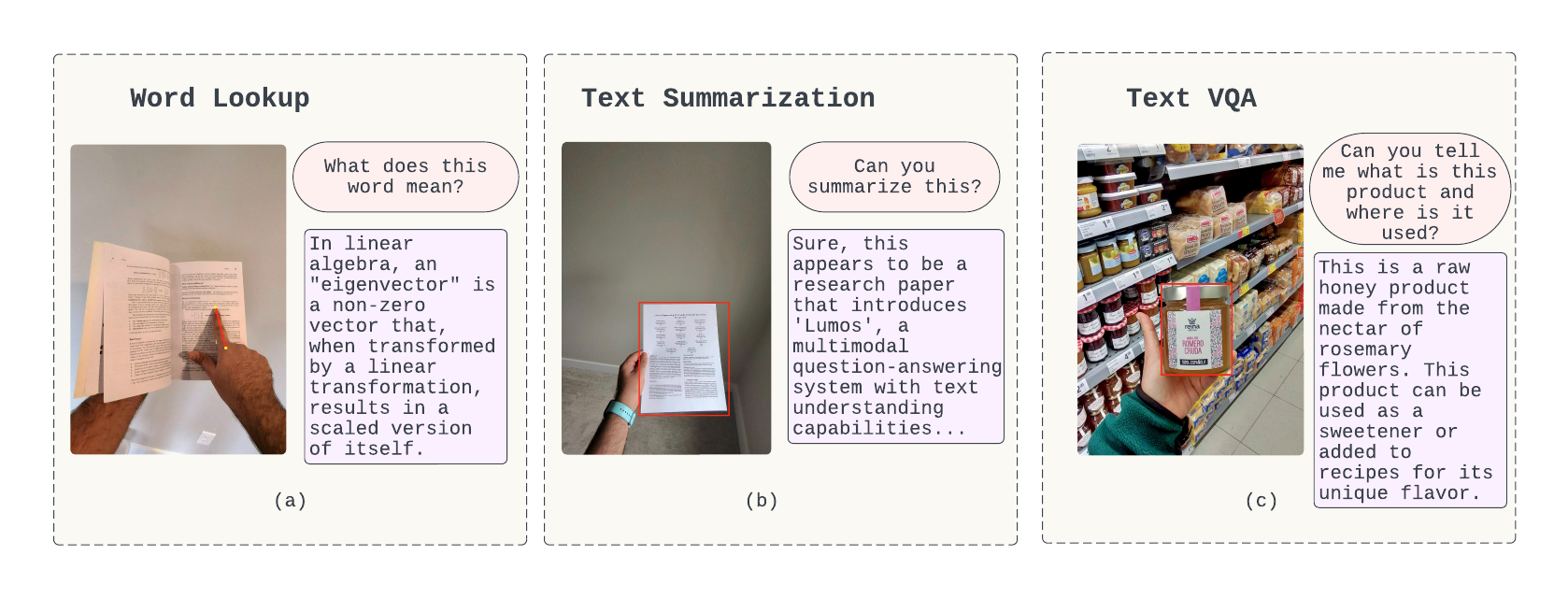}
    \caption{Text based use cases that \textsc{Lumos} supports.}
    \label{usecases}
\end{figure*}

\begin{figure*}[htp]
    \centering
    \begin{subfigure}[b]{0.4\textwidth}
        \centering
        \includegraphics[height=1.0in]{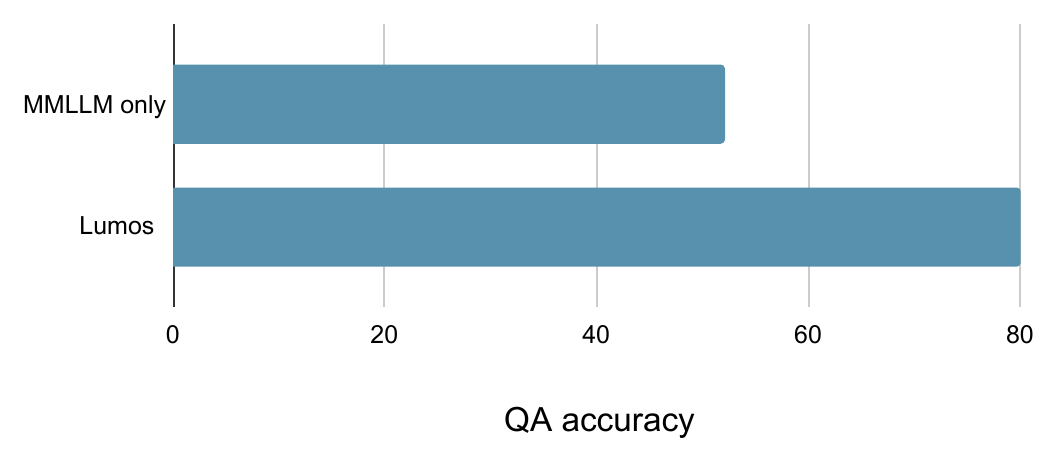}
        \caption{\textsc{Lumos} achieved 80\% QA accuracy, while adding the STR component increased the QA accuracy by 28\%}
        \label{qa_acc}
    \end{subfigure}%
    ~ 
    \begin{subfigure}[b]{0.5\textwidth}
        \centering
        \includegraphics[height=0.9in]{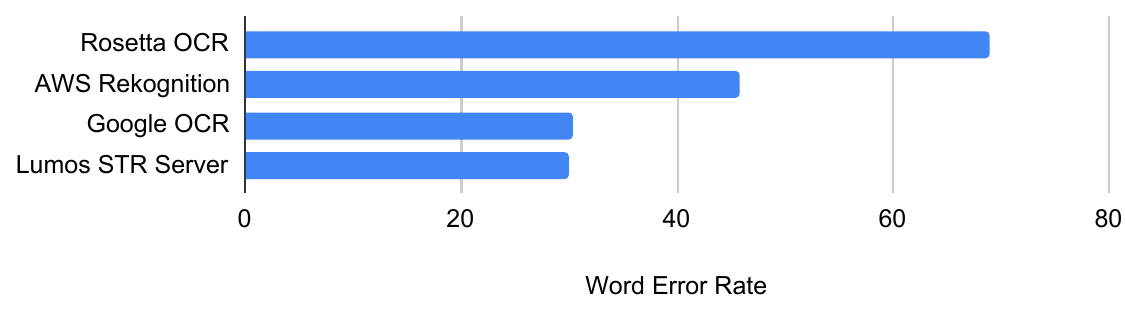}
        \caption{\textsc{Lumnos} STR has lowest word error rate compared with other STR solutions}
        \label{wer_acc}
    \end{subfigure}
    \caption{\textsc{Lumos} Quality metrics}
\end{figure*}

\section{Introduction}
\label{sec:intro}
Visual question answering has been a research area for nearly a decade \cite{VQA15} and has gained increased attention with recent progress on LLMs and vision language pre-training (Multi-Modal LLMs). Industry anticipates that very soon, we will have smart assistants that understand scenes/images just as well as humans \cite{openai2023gpt4, geminiteam2023gemini}. In this paper, we focus on one key abilities needed for scene understanding, visual understanding and question-answering related to text in the scene. We describe \textsc{Lumos}\footnote{Derived from the Latin word 'lumen', meaning 'light'.}, an end-to-end system for multimodal text-understanding. Lumos is well-suited for real-world application and readily leverages on-device processing to enable smooth user experiences.

\smallskip
\noindent
 Figure~\ref{usecases} shows example user interactions for some of \textsc{Lumos}'s use-cases. At the first glance, one may think this problem is already solved by Multimodal Large Language Models (MM-LLMs). In~\cite{openai2023gpt4,geminiteam2023gemini}, MM-LLMs demonstrated capabilities understanding texts from images without a standalone STR component. It would seem straight-forward to enable this feature for edge devices by taking a picture using the on-device camera, transfering to a cloud-based service, along with any voice recognition of user questions, and then having an MM-LLM answer the question using the image. If text recognition is sub-optimal when relying purely on the MM-LLM, one might choose to run a separate Scene Text Recognition (STR), another mature technique, on the image and send the recognized texts to the MM-LLM as prompt input to facilitate text understanding. We will now discuss in detail why such implementations are inadequate and the challenges we solve within \textsc{Lumos}.

\smallskip
\noindent
The first and key challenge we encounter is {\em latency:} just transferring a high-resolution image from device to cloud cost significant time resulting in a poor user experience.  For instance, transmitting an image of size $3k \times 4k$ (standard resolution for today's devices) from a device to the cloud may take several seconds before even running any AI models. And the end-to-end time to get a response would be even longer making for a poor experience. 

\smallskip
\noindent
Alternatively, if we transfer only a low-resolution thumbnail, the transfer time can be significantly reduced (e.g., transferring a thumbnail of size $450 \times 600$ pixels takes only a few hundred ms). However, this results in significantly degraded quality on text recognition. As shown in Table~\ref{tab:ocr_mmllm}, the accuracy of question answering relying solely on running MM-LLM over thumbnails is only 52\%. A separate cloud-based STR can barely recognize texts on the thumbnails either, since the size is too small, illegible even for humans.

\smallskip
\noindent
Now assuming we choose an on-device STR solution, the second challenge is the constrained compute and memory resources on devices. Although running STR models on-device may seem like a viable solution to address latency concerns, current state-of-the-art STR models are not readily suitable for on-device usage; for example, Google's recent work~\cite{gocr2022} features a text detection model that alone has a size of 240MB, impractical for on-device use where several other processes might be running and sharing memory. 

\smallskip
\noindent
The final set of challenges arise with doing STR on in-the-wild text images, which are different from common web images, scanned documents or zoomed-in images. Images taken on-the-go and outdoors can amplify the challenges of STR. 1) The cameras are typically wide angle, and thus the text of interest occupies only a small portion of the image; furthermore, there is often additional background text which can be irrelevant to the user query (see Figure~\ref{usecases}(c)). 2) The text in the scene may not have any uniformity: rotated, diverse orientations and font sizes. 3) The image quality might be poor owing to sub-optimal lighting condition, user movement, and the camera angle. For all of these reasons, traditional OCR (Optical Character Recognition) systems, despite their strong performance on scanned documents and screenshots, can fall short on a STR task in an in-the-wild text setting. As an example, the cloud-based OCR solution Rosetta~\cite{rosetta} exhibits a surprising 53\% Word Error Rate (WER) on our in-the-wild text STR benchmark (see Section~\ref{sec:eval} for details).

\smallskip
\noindent
In this paper, we discuss our results overcoming these three challenges. (1) In our tests, our proposed system has an average end-to-end latency of $\le 5$ seconds, including photo capture, image transfer, on-device STR execution, and on-cloud MM-LLM execution. (2) Our on-device STR models have a total size of $\le 8$Mb, a peak memory footprint of $\le 200$Mb, an average latency of $\le 1$sec, and 0.4 mWh power usage. (3) Despite the low cost, our STR solution achieves competitive quality on public STR benchmarks when compared to state-of-the-art STR solutions from other cloud service providers (Figure~\ref{wer_acc}). On our own in-the-wild text benchmarks, it achieves a $14.6\%$ WER and enables an average accuracy of 80\% on complex text-based QA tasks, improving over vanilla MM-LLM solution by 28\% (see Figure~\ref{qa_acc}).

\smallskip
\noindent
There are three key innovations in \textsc{Lumos}: First, a hybrid approach to multimodal text-understanding with {\em an architecture leveraging components across on-device and on-cloud}. In particular, we conducted on-device STR, such that we can achieve high-quality text recognition results on the full-resolution image; we then send the recognized texts, together with the low-resolution image to the MM-LLM on cloud for question answering; as we run STR in parallel to image transfer, which is the main latency bottleneck, the on-device STR does not add additional latency most of the time (see Section~\ref{sec:arch}). Running STR on the full-resolution image can still be computationally expensive on device, hence our second innovation is an {\em ROI (Region Of Interest) detection solution} that allows the STR to focus on the area of interest and thus reduce the computational overhead. Our ROI detection solution first effectively detects salient areas in the visual, and then crops the salient area as STR input (see Section~\ref{subsec:roi}). Third, we developed {\em a state-of-the-art on-device and resource-preserving STR model}. We optimized our models to run with hardware acceleration resulting in a smaller memory and compute footprint, and efficient battery usage, with minimum sacrifice on quality (see Section~\ref{subsec:detection}-\ref{sec:ondevice}). 

\smallskip
\noindent
To the best of our knowledge, we are the first to propose a multimodal assistant with text understanding capabilities that heavily leverages on-device computation.  We summarize our key contributions as follows:
\begin{itemize}
    \item We propose \textsc{Lumos}, an end-to-end (E2E) multimodal assistant system with text understanding capabilities; through careful placement of components on-device or on-cloud, we are able to achieve high quality, low latency, and minimize on-device resource usage. 
    \item We present an on-device STR pipeline with a set of models for ROI detection, text detection, text recognition, and reading order reconstruction that together achieved high quality (WER=14.6\%) and low cost (latency=0.9s, peak runtime memory=200 Mb, power=0.4 mwh on testing device). 
    \item Through a comprehensive evaluation of our system on QA benchmarks, we validated the high effectiveness and efficiency of our system.
\end{itemize}

\section{Previous work}
\textbf{OCR and STR.} The field of OCR has been a focal point of research for many years. However, the spectrum of difficulty in recognizing text in natural environments is notably broad. At one end, OCR's application to scanned documents containing well-structured printed text is widely recognized as one of the most successful implementations of computer vision~\cite{docvqa_21, icdar_19}. Conversely, STR focuses on recognizing text in the wild, which still represent a significant challenge due to the larger variance of wild text objects~\cite{rosetta, wang_belongie_10, jaderberg_16, shi_robust_str_16, stride_21, paddle_ocr_17}. The STR problem we are solving in this paper considers in-the-wild text images (so the area of interest is considerably smaller), and needs to be tackled on device, thus is much harder and requires better model designs and tuning. 

\smallskip
\noindent
\textbf{On-device STR.} When it comes to STR on-device, in \cite{ppocr2020} an extremely lightweight OCR system with a size of only 3.5Mb is proposed; the model achieves impressive latency on GPUs but still falls short when it comes to CPUs. Munjal et al.~\cite{stride_21} describes an efficient lightweight STR system, which has only 0.88M parameters and performs real-time text recognition at a speed of 2.44 ms per word crop of size $16*64$. In comparison, the STR solution described in this paper takes 0.29 ms per word crop of size $48 * 320$. 

\smallskip
\noindent
\textbf{Multimodal LLMs and Text Recognition Ability} More recently, MM-LLMs have demonstrated potential in addressing a variety of tasks, including text recognition~\cite{geminiteam2023gemini, alayrac2022flamingo, feng2023unidoc, ye2023ureader, zhu2023minigpt4, openai2023gpt4, liu2023visual}. While the current trend leans towards the use of all-modality LLMs, they have limitations particularly in handling text-in-the-wild scenarios. Furthermore, the challenges associated with high transfer latency as described in Section~\ref{sec:intro} makes these models impractical for immediate use \cite{liu2023hidden, shi2023exploring}. A different approach, the {\em Flamingo} models~\cite{alayrac2022flamingo, awadalla2023openflamingo}, have shown impressive performance on tasks such as generic VQA and captioning, but fall short when compared to \cite{hu2023bliva} on text rich VQA. Both sets of models are sub-optimal compared to OCR-assisted VQA as we discussed in this paper and are not optimized for memory and compute at inference time.

\section{Overall Architecture}
\label{sec:arch}
\begin{figure*}[htp]
    \centering
    \includegraphics[width=0.85\textwidth]{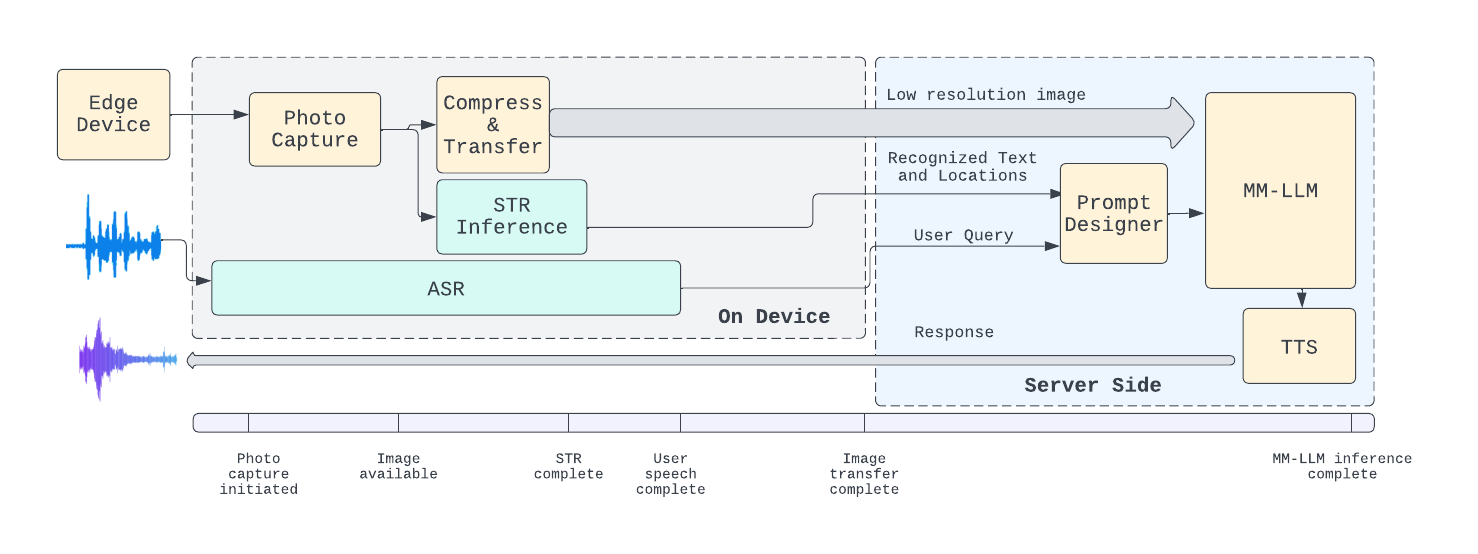}
    \caption{Overall architecture of \textsc{Lumos}. The width of the blocks on device are roughly represents runtime latency. The arrow width roughly represents to the size of the payload being transferred. Blue blocks indicate models using hardware acceleration.}
    \label{arch}
\end{figure*}

We now describe the overall architecture of \textsc{Lumos} (see Figure~\ref{arch}). To simplify, we focus on multimodal use cases, assuming a picture will be taken once the user triggers the flow, and the device will provide the image at two resolutions $3K\times 4K$ (full resolution), and $450\times600$ (thumbnail).

\smallskip
\noindent
\textbf{Device-side:} At the device side, when a user gives a voice query, three components will start in parallel. First, Automatic Speech Recognition (ASR) starts processing the query after a wake word. Second, 
the photo capture, compression (e.g., from a $3k\times4k$ full-resolution image to a  $450\times600$ thumbnail) and transfer to cloud will begin in parallel to the voice query completion (to reduce overall system latency). Third, the STR component will start as soon as the full-resolution image is ready. As indicated by in Figure~\ref{arch},  we carefully design the system to parallelize the time consuming components, STR inference and image transfer, to reduce latency.

\smallskip
\noindent
\textbf{Cloud-side:} The cloud side hosts a {\em MM-LLM model}, which takes as input the low-resolution thumbnail, a prompt composed of the recognized texts and their coordinates from STR, and the user query from ASR, and generates the answer response. An illustrative prompt to MM-LLM can be found in Appendix Table \ref{tab:prompt}. Subsequently, the {\bf TTS (Text-to-Speech)} component translates the response to voice signal and sends back to the user.

\smallskip
This architecture incorporates three design choices we have carefully made.
\begin{itemize}
    \item {\em Where to do STR?} As discussed in detail in Section~\ref{sec:intro}, to reduce latency, we transfer {\em only} a low-resolution image to the cloud. However, neither an MM-LLM nor an STR model can achieve desired quality on such a low-resolution image, especially given that the text area is typically quite small in the in-the-wild text image. We thus apply STR on device with the full-resolution image, and only on the region of interest (see section \ref{subsec:roi} for details).
    \item {\em How to cut the STR latency?} Running STR on device can be time-consuming. To reduce this latency, we took two actions: 1) use hardware acceleration (see section \ref{sec:ondevice}), 2) execute STR and image transfer in parallel (see Figure~\ref{arch}). With this design, for the majority of the cases STR does not add extra latency.
    \item {\em How to extend to MM-LLM use cases where STR is not necessary to answer the user question?} Ideally, we wish to build a {\em single} multimodal assistant that can answer text-heavy questions as well as generic questions where text comprehension is not necessary. Determining whether a user question is based on the text in the scene requires an NLU (Natural Language Understanding) component, which can take extra time and may have limited quality with the limited computation power on device. We instead conduct STR in all cases and defer the decision to the MM-LLM on the cloud. This approach is feasible only because of our significant reduction of latency (particularly through parallelization) and optimization of hardware efficiency for STR.
\end{itemize}

It is worth mentioning that placing STR on-device poses significant constraints on the model's architecture, latency, memory, and battery consumption, in addition to the quality challenges for  in-the-wild text STR discussed in Section~\ref{sec:intro}. Despite these limitations, our on-device STR model achieves strong performance compared to three state-of-the-art cloud STR solutions that do not have such constraints (see Table~\ref{tab:externalcompare} for details). In the next section, we describe how we achieve this.

\begin{figure*}[htp]
    \centering
    \includegraphics[width=0.9\textwidth]{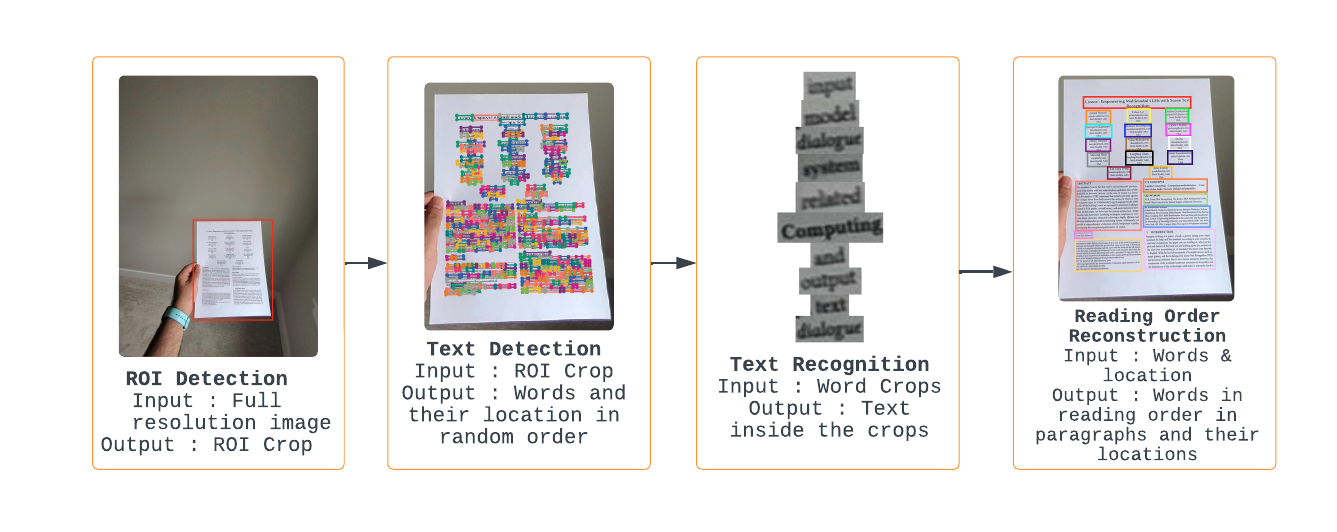}
    \vspace{-0.5cm}
    \caption{On-device STR component flow of \textsc{Lumos}.}
    \vspace{-0.3cm}
    \label{on_device_str}
\end{figure*}

\section{Scene-Text Recognition}
\label{sec:OCR}

We now describe our core technique---the on-device STR. This pipeline contains four sub-components as depicted in Figure~\ref{on_device_str}.
\begin{itemize}
    \item \textbf{Region of Interest (ROI) detection} takes an image as input (at both $3k\times4k$ resolution and a thumbnail resolution), outputs a cropped image (about $1k\times1.3k$ size) that contains all the text likely needed to answer the user query. This component plays a key role to ensure that we run the rest of the STR pipeline only on the relevant portion of the input image, reducing both computational cost and background noise.
    \item \textbf{Text detection} takes the cropped image from ROI detection as input, detects words, and outputs the identified bounding box coordinates for each word.
    \item \textbf{Text recognition} takes the cropped image from ROI detection and the word bounding box coordinates from Text detection as input, returns the recognized words.
    \item \textbf{Reading-order reconstruction} organizes recognized words into paragraphs and in reading order within each paragraph based on the layout. It outputs text paragraphs as well as their location coordinates.
\end{itemize}
We note that in most previous works STR refers to only the Text detection and Text recognition parts. We included two additional components---ROI detection and Reading order reconstruction---in our STR system to address \textsc{Lumos} specific challenges. The primary challenges we face include the limited hardware for inference and the large variation of texts in the wild. We address these challengees through {\em careful model architecture selection} and {\em training data curation and augmentation}, as we discuss in detail next.

\subsection{ROI Detection}
\label{subsec:roi}
{\bf Motivation}
ROI detection plays a key role for on-device STR and there are three motivations behind it. First, as shown in Figure~\ref{usecases}(b), because of the nature of in-the-wild text images, the text area of interest often occupies only a small fraction of the image, even if the object is only an arm length away from the device. Running STR directly on the full-resolution image can be prohibitively expensive with the limited computational power of the device, whereas downsizing the image can make the texts too small to be legible even to humans. Second, as shown in Figure~\ref{usecases}(c), the image may contain a lot of background text that are irrelevant to the user query, such as text from products on the shelves. Recognizing these texts consumes the limited hardware resources, increases the latency, and confuses the MM-LLM at the downstream. Third, users often hold the paper or the object of interest like in Figure~\ref{usecases}(c), or point to the particular words or phrases like in Figure~\ref{usecases}(a), where those gestures provide critical clues for ROI detection. These motivations underscore the importance of identifying the ROI before proceeding with other steps in STR.

\smallskip
\noindent
{\bf Problem definition and challenges} The ROI detection module uses a low resolution thumbnail $450\times600$ to detect the ROI, and returns the cropped area from the raw image $3k\times4k$ containing the ROI. A major challenge for ROI is the non-holding or non-pointing hands in the picture, which can lead to wrong detection results (see example in Figure~\ref{fig:fp_hand} in the Appendix).

\smallskip
\noindent
{\bf Solution and modeling}
We treat ROI detection as an object (salient area) detection problem, facilitated with keypoint detection in presence of {\em a pointing finger}. For finger pointing, we detect two key points---the last joint and the tip of index finger; the two points formulate a pointing vector, as shown in Figure \ref{usecases}(a). We train a model that jointly detects both the ROI and the two keypoints (when present). If the keypoints are detected, we include an additional prompt to the downstream MM-LLM, describing the pointing event as well as the words and the paragraphs closest to the tip of the index finger in the direction of the pointing vector. 
We use the Mask-rcnn~\cite{maskrcnn2018} model since it can provide a unified framework for both object and keypoint detection. We apply inference on the $450\times600$ thumbnail. 

\smallskip
\noindent
{\bf Training data} We trained the model using 80K in-the-wild text images annotated with salient regions, and 20K images with hand holding or finger pointing. To reduce false positives caused by accidental hands, we included 10K images with a hand that is neither holding nor pointing as hard negatives in our training data.

\subsection{Text Detection}
\label{subsec:detection}
{\bf Problem definition and challenges} Text detection takes the cropped image in full-resolution as input, predicts location of each word as bounding boxes.
There are three challenges for detecting text in the wild: C1. the text size can be very small (e.g., ingredients on a coke can at arm length) or very big (e.g., storefront); C2. text can often be tilted with the nature of the image; C3. we are not able to use state-of-the-art text detection model architectures like~\cite{mts2019, gocr2022} with the on-device constraint. 

\smallskip
\noindent
{\bf Solution and modeling} 
To account for the tilted text (C2), our detector predicts rotated bounding box as mentioned in \cite{rrpn2018}. To be computationally efficient (C3), we use an anchor-free single-stage detector as described in~\cite{fcos2019} (instead of a two-stage detector). We use FBNetv2 (with 1.1 million parameters)~\cite{Wan2020FBNetV2DN} with PAN neck~\cite{pan2018} for the backbone of the detector. 
FBNetv2 is a CNN model designed for transforming input images into feature maps; this backbone not only is computationally efficient (C3) but also provides strong image features at different scales (C1).
For the loss, we use a variant of the well-known focal loss \cite{varifocal2021} as classification loss, and the KLD loss~\cite{kld2022} as our box regression loss for its state-of-the-art performance on rotated box (C2).

\smallskip
\noindent
{\bf Training data} 
Our training data consist of 140K images with 6 million annotated bounding boxes, combining
public STR datasets like text OCR~\cite{textocr2021} and in-house annotations on in-the-wild text images.
To address the challenge of text scale variation (C1), we applied aggressive {\em scale jittering}, data augmentation that increases or reduces input image sizes, to create variational sizes of bounding boxes in training data.

\subsection{Text Recognition}
\label{subsec:recognition}
{\bf Problem definition and challenges} Text recognition takes the image crop from ROI detection and the word bounding box coordinates, and outputs the recognized words for each box. There are three key challenges we need to address: C1. huge diversity in the widths of bounding boxes (e.g., URLs tend to be longer, price tags tend to be extremely small); C2. diversity of text appearances in terms of font, size, orientation, and background; C3. existence of (quite some) text detection errors; C4. hardware constraints.

\smallskip
\noindent
{\bf Solution and modeling} 
We transform the problem of recognizing a word into the problem of recognizing a sequence of characters. 
Because of hardware acceleration constraints (C4) as we will describe in Section~\ref{sec:ondevice}, we are limited to using fixed width and height for each bounding box. Therefore, we scale each bounding box to a fixed height of 48 pixels and a fixed width of 320 pixels to ensure that the input to the model is consistent and can be processed efficiently. Based on statistics we assume that each individual character has a width of 8 pixels. Thus, we recognize a maximum of 40 characters ($320/8$) per bounding box; a word rarely exceeds this limit. The final recognizer output is a posterior of shape 40 x $|alphabets|$ and the size of the alphabets in our model is top-150 most frequently used Latin characters obtained from the training data. 

We again use the FBNetv2 backbone and train the model using CTC (Connectionist Temporal Classification) loss, as it can handle variable-length input sequences (C1) and has lower latency and computational complexity (C4), critical in dense text scenarios.

\smallskip
\noindent
{\bf Training data} 
During training, to handle the extreme variations in bounding box lengths (C1), we employ {\em curriculum learning}; that is, we gradually increase the complexity of the input images. We begin with words containing a maximum of 16 characters and progressively increase the character limit up to a maximum of 40 characters. This helps the model learn the necessary features and patterns more effectively.

Overall, the recognizer model is trained on ~3M word bounding boxes, with 15\% being synthetically generated to increase the robustness of the model. 
To be more robust against detector errors (C3), we introduce random cropping around the boundaries of the bounding boxes based on error patterns we have observed in detector evaluation, combined with jittering. We incorporated RandAug~\cite{randaug}, which applies random combinations of image transformations such as rotation, shearing, brightness adjustment, and contrast adjustment to input images. By exposing the model to a wide range of transformed images, it learns to be more robust to these transformations and generalizes better to new, unseen data (C2).

\subsection{Reading Order Reconstruction}
\label{subsec:ROR}
\begin{figure}
    \includegraphics[width=0.35\textwidth]{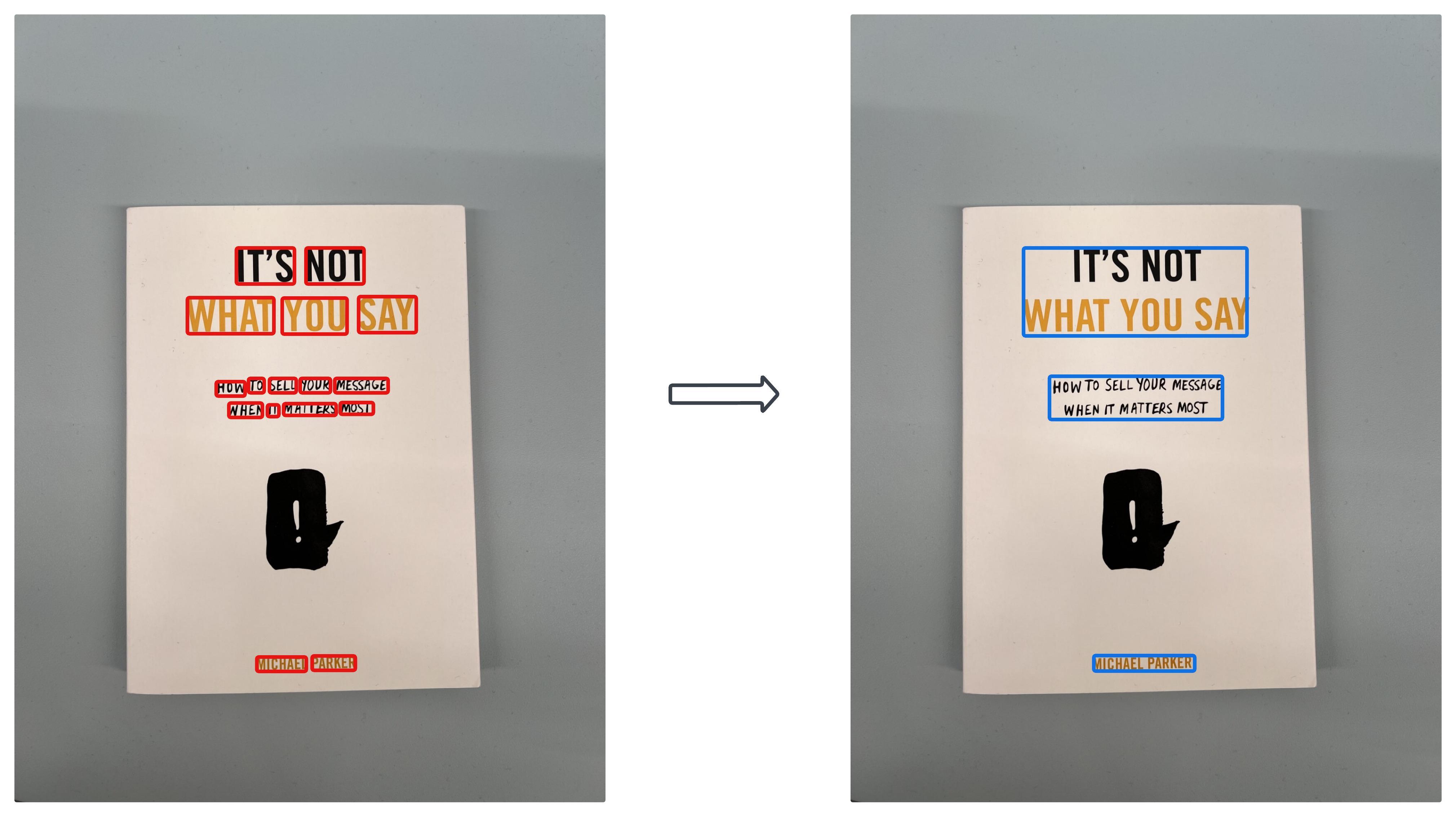}
    \caption{Left: Word bounding boxes. Right: Paragraphs from out Reading Order Reconstruction component}
    \label{grouping}
\end{figure}

{\bf Problem definition} The Reading Order Reconstruction module connects the words to paragraphs, returns the words in the paragraph in reading order, together with the coordinates of each paragraph. Figure \ref{grouping} shows sample paragraphs. 

\smallskip
\noindent
{\bf Solutions} We identify paragraphs in three steps. First, we connect the words to paragraphs. We expand the word bounding boxes both vertically and horizontally by predefined ratios, as shown in Figure~\ref{expandbox}. The expansion ratios are selected to fill the gaps between words within a line and lines within a paragraph and are the same for all bounding boxes.
We then group bounding boxes that have significant overlap after expansion as a paragraph. 
For each paragraph, we then apply {\em raster scan} (i.e., sorting by Y coordinate then X) to the words to generate the paragraph in reading order.
Finally, we compute the location of the paragraph by finding the minimum area rectangle enclosing all words in the paragraph. See Algorithm \ref{grouping_alg} in the Appendix for detailed description of the Reading order reconstruction module.

We found this simple heuristic approach achieves a good quality most of the time with low computation cost. The accuracy for this module is 92\% using metrics defined in~\cite{disgo2023}.

\section{On-Device Export}
\label{sec:ondevice}
As mentioned in the introduction, \textsc{Lumos} need to be compatible with devices to make our smart assistant more accessible. We evaluated our on-device system’s performance with on our testing devices, which is equipped with hardware accelerators for deep learning models. We describe the process of exporting our models to the testing device as well as the memory/latency in this setting.

\begin{figure}
    \includegraphics[width=0.425\textwidth]{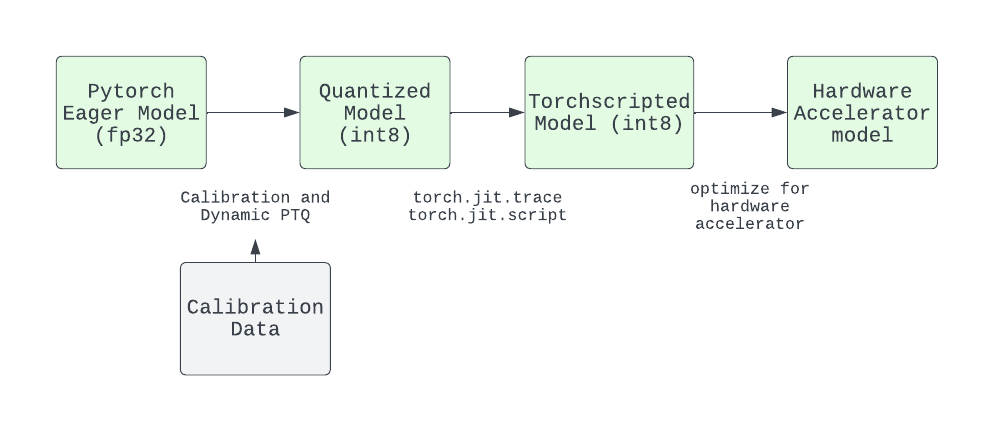}
    \caption{Model Export Pipeline}
    \label{ondevice}
\end{figure}
\begin{enumerate}
    \item {\bf Quantization to int8} We first quantize the float32 models to int8 models to save inference latency and runtime memory. We use Post Training Quantization (PTQ)~\cite{white2021} to do this, because the sizes of our models were relatively small and PTQ requires a calibration step only after models are full trained.
    \item {\bf On-device CPU models} We next transfer the models to TorchScript models using packages provided by PyTorch. This provides a model that is executable on CPU of the device.
    \item{\bf On-device hardware accelerated models} Modern devices often comes with a hardware accelerator for deep learning models. To utilize this, we take a further step making our model hardware accelerator compatible, and evaluate the latency of our system on hardware accelerator.
\end{enumerate}

We emphasize that the model execution efficiency is achieved with cost. First, we are constrained to use quantization and hardware accelerator friendly models, limited our modeling choices as stated in Section~\ref{sec:OCR}.  Second, quantization and model export would cause accuracy drops in the ML models. Regardless, our system still achieves competitive performance compared to other STR services as we show soon in Section~\ref{sec:eval}.

\section{Experimental Results}
\label{sec:eval}
We answer two questions with our experiments: 1) How good is \textsc{Lumos} as an end-to-end text visual question answering system? 2) What is the quality, efficiency, and hardware usage for our on-device STR solution?

\begin{table}
\small
\centering
\caption{Evaluation dataset details}
\label{tab:datasetdes}
\begin{tabular}{lll}
\toprule
Name & Size \\
\midrule
In-house wild text benchmark &  968 images, 47K words \\
Public  wild text benchmark &  1.7K images, 146K words \\
\midrule
Summarization question set & 245 images, 735 questions \\
Word Lookup question set & 200 images, 600 questions \\
Other question set & 200 images, 600 questions \\
\bottomrule
\end{tabular}
\end{table}
\subsection{Experiment Setup}
\textbf{Datasets} Table~\ref{tab:datasetdes} summarizes the datasets we used for evaluation. We have two benchmarks: {\em In-house wild text benchmark} and {\em Public wild text benchmark} \cite{heirtext}. {\em In-house wild text benchmark} contains 968 in-the-wild text images taken from an edge device and contains 47K word boxes. The benchmark contains annotations for the word boxes and transcriptions, and in addition annotations for salient areas for ROI evaluation. {\em Public wild text benchmark} is a broadly-used STR benchmark, containing 1.7K images and 146K word boxes.

\smallskip
\noindent
We then created task-specific datasets to evaluate end-to-end quality of summarization, word lookup and a few other tasks on the {\em In-house wild text benchmark}. We first sampled text-heavy images from the benchmark, and then our annotators created $\sim 3$ task-related questions for each image. 

\smallskip
\noindent
\textbf{Metrics definition} We have two major metrics. To understand the end-to-end question answering quality, we measure {\em QA accuracy} as the percentage of successful responses among all answers. A group of raters manually decided the correctness of each response judging from the image, user query and generated response, based on the relevancy, fluency and factual accuracy of the response.

\smallskip
\noindent
To understand the quality of STR solutions, we measured the {\em Word Error Rate} (WER), a standard metric extensively used in the domain of speech \cite{shenoy21_interspeech, shenoy-etal-2021-asr, bekal_21} and text recognition \cite{dingliwal22_interspeech, dingliwal_wecnlp_21}. WER considers 3 types of errors: 1) Deletion: a ground truth word that is not detected; 2) Insertion: a prediction that is not matched to any ground truth word box; 3) Substitution: a prediction that matches a ground truth box, but the word recognized is different from the ground truth. WER is the sum of Deletion, Insertion, Substitution errors divided by the total number of words in the ground truth. With the existence of insertion errors, WER can be higher than 1. A lower WER is indicative of higher quality of the models.

\begin{table}[t!]
\small
\centering
\caption{QA accuracy of \textsc{Lumos} variants on task-specific benchmarks. On-device STR boosts QA accuracy by 28\%.}
\label{tab:ocr_mmllm}
\begin{tabular}{lllll}
\toprule
System & Summarization & Word Lookup & Others & Avg\\
\midrule
MMLLM only & 53.0\% & 43.0\% & 60.1\% & 52.0\%\\
 + STR & 87.7\%   & 65.0\% & 81.3\% & 78.0\% \\
 + STR + Pos  & \textbf{88.3\%}   & \textbf{67.2\%} & \textbf{83.3\%} & \textbf{79.6\%} 
\\
\bottomrule
\end{tabular}
\end{table}
\subsection{End-to-End Quality}
We evaluated the overall quality of three variants of \textsc{Lumos}: 1) MMLLM only: we provide only the $450\times600$ thumbnail and user query to the MM-LLM; 2) MM-LLM+STR: we in addition provide the text output from the on-device STR to MM-LLM; 3) MM-LLM+STR+Positions: we in addition provide the paragraph location (from reading order reconstruction module). See Table~\ref{tab:prompt} for detailed input formats of these variants. 

\smallskip
\noindent
Table~\ref{tab:ocr_mmllm} compares the QA accuracy of the three variants on the task-specific E2E datasets.
We have four observations. First, \textsc{Lumos} obtains a high average QA accuracy, 80\%, in question answering. Second, the on-device STR significantly improves QA accuracy on all three tasks over MM-LLM only (80\% vs. 52\%) . The improvement is particularly large for the summarization task (+35\%), where \textsc{Lumos} needs to comprehend dense texts. Third, sending positions to MM-LLM further improves the performance on all tasks (+1.6\%), as it allows the model to better handle the spatial relationships between words in the scene. Finally, among different tasks, we observe the best quality on summarization (88\%), which has higher tolerance on small recognition errors; the quality on word lookup is lowest (67\%), as we observe a large variety of hand-word positions, making the problem much more difficult.

\begin{table}
\small
\centering
\caption{WER comparison on public wild text benchmarks. \textsc{Lumos} STR obtains the lowest WER with a small size, and the on-device model sacrifices quality only slightly.}
\label{tab:externalcompare}
\begin{tabular}{llllll}
\toprule
Model& WER & Del & Ins & Sub& \#Params \\
\midrule
\textit{Public wild text benchmark \cite{heirtext}} & & & & &\\
\midrule
Rosetta OCR & 68.9\%& 58.1\%& 2.3\% & 8.5\% & ~15Mb\\
AWS Rekognition~\cite{aws_rekognition} & 45.8\%& 38.1\% & 1.6\% & 6.1\% & -\\
Google OCR~\cite{google_ocr} & 30.4\% &9.4\% &9.5\% & 11.5\% & 240Mb+\footnotemark \\
\textbf{\textsc{Lumos} STR Server} & \textbf{29.9\%}&\textbf{17.7\%}&\textbf{2.5\%}&\textbf{9.7\%} & ~30Mb\\
\textsc{Lumos} STR Device & 32.4\%&18.5\%&2.7\%&11.2\%& \textbf{~8Mb} \\
\midrule
\textit{In-house wild text benchmark} & & & & &\\
\midrule
Rosetta OCR & 53\%& 46.0\%& 1.1\% & 5.9\% & ~15Mb\\
\textbf{\textsc{Lumos} STR Server} & \textbf{13\%}&\textbf{4.7\%}&\textbf{1.4\%}&\textbf{6.9\%} & ~30Mb\\
\textsc{Lumos} STR Device & 14.6\%&5.1\%&1.8\%&7.7\%& \textbf{~8Mb} \\
\bottomrule
\end{tabular}
\end{table}

\footnotetext[2]{Estimated based on \cite{gocr2022}, using the size of MaX-DeepLab-S \cite{maxdeeplab21}}

\subsection{STR quality}
{\bf \textsc{Lumos} STR quality}
We next compare quality of 5 STR Systems: 1) Rosetta~\cite{rosetta}, a well known STR system from the research community; 2) Google Cloud OCR~\cite{google_ocr}; 3) AWS Rekognition~\cite{aws_rekognition}; 4) \textsc{Lumos} STR Cloud: \textsc{Lumos} STR running on cloud; 5) \textsc{Lumos} STR Device: \textsc{Lumos} STR running on our device hardware. For a fair comparison, we removed punctuations from the benchmarks since different baseline STR systems treat them differently, as a separate word or part of a word. We also removed words smaller than 8 pixels high since it is hard for humans to read. 

\smallskip
\noindent
Table~\ref{tab:externalcompare} shows the WER of each solution, together with error breakdowns in terms of deletion, insertion, substitution errors. We have four observations. 1) \textsc{Lumos} STR has a reasonably low WER, 30\% on the public benchmark and 13\% on the in-house benchmark. 2) \textsc{Lumos} STR outperforms Rosetta, AWS, and Google, despite never trained on the public wild text benchmark (we do not know if Google and AWS were trained on the public wild text benchmark). Rosetta made a lot of deletion errors as it missed small texts and has a low word limit per image. Similarly, AWS has a low word limit per image, leading to high deletion errors. 3) \textsc{Lumos} STR Device is smallest in model size with only $\sim~$8Mb parameters; nevertheless, it sacrifices WER by only 1-2\% comparing with the on-server model and still has a competitive performance. 4) Finally, among different types of errors, Substitution errors is only a small portion (<10\%), showing that word detection is a much bigger challenge than word recognition for STR tasks. 

\begin{table*}[htp]
\centering
\caption{WER gains from each component}
\label{tab:attribution}
\begin{tabular}{llll}
\toprule
Component& Reason & WER & Comp. to baseline\\
\midrule
Baseline (Rosetta OCR) & - & 53\% \\
+ROI detection & avoid aggressive input image downsizing& 42\% & -11\% \\
+Text Detection & stronger model, data augmentation, &  26\% & -16\%\\
&more in domain training data, increased word limit &   \\
+Text Recognition & synthetic data on rare/hard symbols & 13\% & -13\%\\
&det error simulation, RandAug& \\
+on-device export & model quantization error & 14.6\% & +1.6\%\\
\bottomrule
\end{tabular}
\end{table*}

\smallskip
\noindent
{\bf Ablation study} We now listed the main drivers for WER improvements. We compared with Rosetta, a two-step STR system (faster-rcnn \cite{faster2016} word detector and CNN + CTC recognizer) on the {\em In-house wild text benchmark}. There are three contributors for quality improvements as shown in Table~\ref{tab:attribution}. 
\begin{itemize}
  \item ROI detection allows us to run our detection and recognition on a text-dense cropped region in original size, instead of on an aggressively downsized (3x-4x) full image, thus reducing WER by 11\%, and especially reducing WER on small-font texts. 
  \item Our detection model uses additional in-domain data and data augmentation for training to improve robustness, and increases word limit per image, thus reducing WER by 16\%. In particular, we increased recall of detecting word boxes, thus reducing deletion errors, in detection of small text (<15 pixels tall) by 14\% and of large text (>120 pixels tall) by 20\%.
  \item Our recognition model used data augmentation to accommodate more variations for text in the wild, thus reducing WER by 13\%. 
\end{itemize}

Finally, these improvements are well preserved in model quantization and export, which increased WER by only 1.6\% but achieved huge efficiency gains as we discuss soon in Section~\ref{subsec:cost}.

\begin{table}[t!]
\centering
\caption{Recall for ROI detection. On average our ROI method is able to reduce image size by 25\% while including 99\% words of interest.}
\label{tab:roirecall}
\begin{tabular}{lll}
\toprule
Method & Recall & Improvement \\
\midrule
Center Crop & 65.9\% & \\
ROI detection & 97.7\% & +31.8\%\\
ROI detection with Hand cues &  99.0\% & +1.3\%\\
\bottomrule
\end{tabular}
\end{table}

\smallskip
\noindent
{\bf ROI detection recall}
To illustrate the effectiveness of the ROI detection component, we compared the performance of 3 image cropping methods: 1) Center Crop: heuristic-rule baseline that crops the 1500*2000 center region (similar as the ROI output size); 2) ROI detection: use an object detection model to detect the region; 3) ROI detection with hand cues: use object detection together with the holding and pointing gestures to detect the region.

We measured ROI quality by word-level {\em recall}---how many words of interest are included in the ROI output region. Table \ref{tab:roirecall} shows the results on the {\em in house wild text benchmark}. We are able to reach 99\% recall with our ROI detection component while reducing image size by 25\% on average. Our model achieves much higher recall (+32\%) than the Center Crop baseline, and including hand cues further improves the recall (+1.3\%).

\begin{table}[H]
\small
\centering
\caption{Model execution metrics. Running the models on hardware accelerator (HA) saved latency by 9X and energy by 3X comparing with running on CPU.}
\label{tab:htpcpu}
\begin{tabular}{llll}
\toprule
Metrics& CPU &  HA & Saving \\
\midrule
Overall on device latency (100 words) & 8390ms & 940ms & 8.9X\\
Text Detection latency & 750ms & 66ms & 11.4X\\
Text Recognition latency & 238ms & 29ms & 8.2X\\
ROI detection latency &   300ms & 30ms & 10X\\
\midrule
Model size & - & 8Mb & -\\
Peak memory footprint & - & 200Mb & -\\
\midrule
Overall on device energy cost & 1.1mwh & 0.4mwh & 2.8X\\
\bottomrule
\end{tabular}
\vspace{-1.0em}
\end{table}
\subsection{STR Efficiency}
\label{subsec:cost}
Finally, we show the efficiency of our STR models in Table~\ref{tab:htpcpu} when running on testing devices. The model export steps generated on-device compatible models with the total size around 8Mb. Running the models on hardware accelerator provided huge gain in terms of both latency (9x) and battery usage (3x).

\section{Conclusion}
This paper presented \textsc{Lumos}, one of the first smart multimodal assistant with strong text understanding
capabilities which is also device compatible. Our comprehensive evaluation demonstrates the effectiveness of our proposed method, outperforming existing approaches in terms of accuracy. Additionally, we have shown that our system meets the stringent latency, size, memory, power, and compute requirements for on-device deployment. Overall, our work represents a significant step towards enabling MM-LLMs to read in real-world scenarios, paving the way for more advanced applications in the fields of computer vision and natural language processing.
Future work includes further optimizations to our on-device models, and research on end-to-end text recognition and visual translation with multimodal large language models.

\begin{acks}
The authors would like to thank Mei-Yuh Hwang, Praveen Krishnan, Guan Pang, Becka Silvert, Renato Sanchez, Crystal Nakatsu, Lucas Kabela, Frank Seide, Samyak Datta, Peyman Heidari, Shashank Jain, Nish Gupta, Kate Ovchinnikova, Rongzhou Shen, Saumya Mukul, Shane Moon, David Strauss, Lintao Cui, Sofiane Djeffal, Megha Tiwari, Vitaly Berov, Shanying Luo for their valuable inputs and contributions.
\end{acks}

\bibliographystyle{ACM-Reference-Format}
\bibliography{main}


\begin{thebibliography}{43}

%%% the Web address.

\ifx \showCODEN    \undefined \def \showCODEN     #1{\unskip}     \fi
\ifx \showDOI      \undefined \def \showDOI       #1{#1}\fi
\ifx \showISBNx    \undefined \def \showISBNx     #1{\unskip}     \fi
\ifx \showISBNxiii \undefined \def \showISBNxiii  #1{\unskip}     \fi
\ifx \showISSN     \undefined \def \showISSN      #1{\unskip}     \fi
\ifx \showLCCN     \undefined \def \showLCCN      #1{\unskip}     \fi
\ifx \shownote     \undefined \def \shownote      #1{#1}          \fi
\ifx \showarticletitle \undefined \def \showarticletitle #1{#1}   \fi
\ifx \showURL      \undefined \def \showURL       {\relax}        \fi
% invisible to TeX
\providecommand\bibfield[2]{#2}
\providecommand\bibinfo[2]{#2}
\providecommand\natexlab[1]{#1}
\providecommand\showeprint[2][]{arXiv:#2}

\bibitem[aws({[n.\,d.]})]%
        {aws_rekognition}
 \bibinfo{year}{[n.\,d.]}\natexlab{}.
\newblock \bibinfo{title}{AWS Rekognition}.
\newblock
\newblock
\urldef\tempurl%
\url{https://aws.amazon.com/rekognition/}
\showURL{%
\tempurl}


\bibitem[goo({[n.\,d.]})]%
        {google_ocr}
 \bibinfo{year}{[n.\,d.]}\natexlab{}.
\newblock \bibinfo{title}{Google Cloud OCR}.
\newblock
\newblock
\urldef\tempurl%
\url{https://cloud.google.com/vision/docs/ocr}
\showURL{%
\tempurl}


\bibitem[(2023)(2023)]%
        {openai2023gpt4}
\bibfield{author}{\bibinfo{person}{OpenAI (2023)}.}
  \bibinfo{year}{2023}\natexlab{}.
\newblock \bibinfo{title}{GPT-4 Technical Report}.
\newblock
\newblock
\showeprint[arxiv]{2303.08774}~[cs.CL]


\bibitem[Alayrac et~al\mbox{.}(2022)]%
        {alayrac2022flamingo}
\bibfield{author}{\bibinfo{person}{Jean-Baptiste Alayrac},
  \bibinfo{person}{Jeff Donahue}, \bibinfo{person}{Pauline Luc},
  \bibinfo{person}{Antoine Miech}, \bibinfo{person}{Iain Barr},
  \bibinfo{person}{Yana Hasson}, \bibinfo{person}{Karel Lenc},
  \bibinfo{person}{Arthur Mensch}, \bibinfo{person}{Katie Millican},
  \bibinfo{person}{Malcolm Reynolds}, \bibinfo{person}{Roman Ring},
  \bibinfo{person}{Eliza Rutherford}, \bibinfo{person}{Serkan Cabi},
  \bibinfo{person}{Tengda Han}, \bibinfo{person}{Zhitao Gong},
  \bibinfo{person}{Sina Samangooei}, \bibinfo{person}{Marianne Monteiro},
  \bibinfo{person}{Jacob Menick}, \bibinfo{person}{Sebastian Borgeaud},
  \bibinfo{person}{Andrew Brock}, \bibinfo{person}{Aida Nematzadeh},
  \bibinfo{person}{Sahand Sharifzadeh}, \bibinfo{person}{Mikolaj Binkowski},
  \bibinfo{person}{Ricardo Barreira}, \bibinfo{person}{Oriol Vinyals},
  \bibinfo{person}{Andrew Zisserman}, {and} \bibinfo{person}{Karen Simonyan}.}
  \bibinfo{year}{2022}\natexlab{}.
\newblock \bibinfo{title}{Flamingo: a Visual Language Model for Few-Shot
  Learning}.
\newblock
\newblock
\showeprint[arxiv]{2204.14198}~[cs.CV]


\bibitem[Antol et~al\mbox{.}(2015)]%
        {VQA15}
\bibfield{author}{\bibinfo{person}{Stanislaw Antol}, \bibinfo{person}{Aishwarya
  Agrawal}, \bibinfo{person}{Jiasen Lu}, \bibinfo{person}{Margaret Mitchell},
  \bibinfo{person}{Dhruv Batra}, \bibinfo{person}{C.~Lawrence Zitnick}, {and}
  \bibinfo{person}{Devi Parikh}.} \bibinfo{year}{2015}\natexlab{}.
\newblock \showarticletitle{{VQA:} Visual Question Answering}.
\newblock \bibinfo{journal}{\emph{CoRR}}  \bibinfo{volume}{abs/1505.00468}
  (\bibinfo{year}{2015}).
\newblock
\showeprint[arXiv]{1505.00468}
\urldef\tempurl%
\url{http://arxiv.org/abs/1505.00468}
\showURL{%
\tempurl}


\bibitem[Awadalla et~al\mbox{.}(2023)]%
        {awadalla2023openflamingo}
\bibfield{author}{\bibinfo{person}{Anas Awadalla}, \bibinfo{person}{Irena Gao},
  \bibinfo{person}{Josh Gardner}, \bibinfo{person}{Jack Hessel},
  \bibinfo{person}{Yusuf Hanafy}, \bibinfo{person}{Wanrong Zhu},
  \bibinfo{person}{Kalyani Marathe}, \bibinfo{person}{Yonatan Bitton},
  \bibinfo{person}{Samir Gadre}, \bibinfo{person}{Shiori Sagawa},
  \bibinfo{person}{Jenia Jitsev}, \bibinfo{person}{Simon Kornblith},
  \bibinfo{person}{Pang~Wei Koh}, \bibinfo{person}{Gabriel Ilharco},
  \bibinfo{person}{Mitchell Wortsman}, {and} \bibinfo{person}{Ludwig Schmidt}.}
  \bibinfo{year}{2023}\natexlab{}.
\newblock \bibinfo{title}{OpenFlamingo: An Open-Source Framework for Training
  Large Autoregressive Vision-Language Models}.
\newblock
\newblock
\showeprint[arxiv]{2308.01390}~[cs.CV]


\bibitem[Bekal et~al\mbox{.}(2021)]%
        {bekal_21}
\bibfield{author}{\bibinfo{person}{Dhanush Bekal}, \bibinfo{person}{Ashish
  Shenoy}, \bibinfo{person}{Monica Sunkara}, \bibinfo{person}{Sravan Bodapati},
  {and} \bibinfo{person}{Katrin Kirchhoff}.} \bibinfo{year}{2021}\natexlab{}.
\newblock \showarticletitle{Remember the Context! ASR Slot Error Correction
  Through Memorization}. In \bibinfo{booktitle}{\emph{2021 IEEE Automatic
  Speech Recognition and Understanding Workshop (ASRU)}}.
  \bibinfo{pages}{236--243}.
\newblock
\urldef\tempurl%
\url{https://doi.org/10.1109/ASRU51503.2021.9688109}
\showDOI{\tempurl}


\bibitem[Borisyuk et~al\mbox{.}(2018)]%
        {rosetta}
\bibfield{author}{\bibinfo{person}{Fedor Borisyuk}, \bibinfo{person}{Albert
  Gordo}, {and} \bibinfo{person}{Viswanath Sivakumar}.}
  \bibinfo{year}{2018}\natexlab{}.
\newblock \showarticletitle{Rosetta: Large Scale System for Text Detection and
  Recognition in Images}. In \bibinfo{booktitle}{\emph{Proceedings of the 24th
  ACM SIGKDD International Conference on Knowledge Discovery \& Data Mining}}
  (London, United Kingdom) \emph{(\bibinfo{series}{KDD '18})}.
  \bibinfo{publisher}{Association for Computing Machinery},
  \bibinfo{address}{New York, NY, USA}, \bibinfo{pages}{71–79}.
\newblock
\showISBNx{9781450355520}
\urldef\tempurl%
\url{https://doi.org/10.1145/3219819.3219861}
\showDOI{\tempurl}


\bibitem[Cubuk et~al\mbox{.}(2019)]%
        {randaug}
\bibfield{author}{\bibinfo{person}{Ekin~D. Cubuk}, \bibinfo{person}{Barret
  Zoph}, \bibinfo{person}{Jonathon Shlens}, {and} \bibinfo{person}{Quoc~V.
  Le}.} \bibinfo{year}{2019}\natexlab{}.
\newblock \showarticletitle{RandAugment: Practical data augmentation with no
  separate search}.
\newblock \bibinfo{journal}{\emph{CoRR}}  \bibinfo{volume}{abs/1909.13719}
  (\bibinfo{year}{2019}).
\newblock
\showeprint[arXiv]{1909.13719}
\urldef\tempurl%
\url{http://arxiv.org/abs/1909.13719}
\showURL{%
\tempurl}


\bibitem[Dingliwal et~al\mbox{.}(2021)]%
        {dingliwal_wecnlp_21}
\bibfield{author}{\bibinfo{person}{Saket Dingliwal}, \bibinfo{person}{Ashish
  Shenoy}, \bibinfo{person}{Sravan Bodapati}, \bibinfo{person}{Ankur Gandhe},
  \bibinfo{person}{Ravi~Teja Gadde}, {and} \bibinfo{person}{Katrin Kirchhoff}.}
  \bibinfo{year}{2021}\natexlab{}.
\newblock \showarticletitle{Efficient domain adaptation of language models in
  {ASR} systems using Prompt-tuning}.
\newblock \bibinfo{journal}{\emph{CoRR}}  \bibinfo{volume}{abs/2110.06502}
  (\bibinfo{year}{2021}).
\newblock
\showeprint[arXiv]{2110.06502}
\urldef\tempurl%
\url{https://arxiv.org/abs/2110.06502}
\showURL{%
\tempurl}


\bibitem[Dingliwal et~al\mbox{.}(2022)]%
        {dingliwal22_interspeech}
\bibfield{author}{\bibinfo{person}{Saket Dingliwal}, \bibinfo{person}{Ashish
  Shenoy}, \bibinfo{person}{Sravan Bodapati}, \bibinfo{person}{Ankur Gandhe},
  \bibinfo{person}{Ravi~Teja Gadde}, {and} \bibinfo{person}{Katrin Kirchhoff}.}
  \bibinfo{year}{2022}\natexlab{}.
\newblock \showarticletitle{{Domain Prompts: Towards memory and compute
  efficient domain adaptation of ASR systems}}. In
  \bibinfo{booktitle}{\emph{Proc. Interspeech 2022}}.
  \bibinfo{pages}{684--688}.
\newblock
\showISSN{2308-457X}
\urldef\tempurl%
\url{https://doi.org/10.21437/Interspeech.2022-824}
\showDOI{\tempurl}


\bibitem[Du et~al\mbox{.}(2020)]%
        {ppocr2020}
\bibfield{author}{\bibinfo{person}{Yuning Du}, \bibinfo{person}{Chenxia Li},
  \bibinfo{person}{Ruoyu Guo}, \bibinfo{person}{Xiaoting Yin},
  \bibinfo{person}{Weiwei Liu}, \bibinfo{person}{Jun Zhou},
  \bibinfo{person}{Yifan Bai}, \bibinfo{person}{Zilin Yu},
  \bibinfo{person}{Yehua Yang}, \bibinfo{person}{Qingqing Dang}, {and}
  \bibinfo{person}{Haoshuang Wang}.} \bibinfo{year}{2020}\natexlab{}.
\newblock \bibinfo{title}{PP-OCR: A Practical Ultra Lightweight OCR System}.
\newblock
\newblock
\showeprint[arxiv]{2009.09941}~[cs.CV]


\bibitem[Feng et~al\mbox{.}(2023)]%
        {feng2023unidoc}
\bibfield{author}{\bibinfo{person}{Hao Feng}, \bibinfo{person}{Zijian Wang},
  \bibinfo{person}{Jingqun Tang}, \bibinfo{person}{Jinghui Lu},
  \bibinfo{person}{Wengang Zhou}, \bibinfo{person}{Houqiang Li}, {and}
  \bibinfo{person}{Can Huang}.} \bibinfo{year}{2023}\natexlab{}.
\newblock \bibinfo{title}{UniDoc: A Universal Large Multimodal Model for
  Simultaneous Text Detection, Recognition, Spotting and Understanding}.
\newblock
\newblock
\showeprint[arxiv]{2308.11592}~[cs.AI]


\bibitem[He et~al\mbox{.}(2018)]%
        {maskrcnn2018}
\bibfield{author}{\bibinfo{person}{Kaiming He}, \bibinfo{person}{Georgia
  Gkioxari}, \bibinfo{person}{Piotr Dollár}, {and} \bibinfo{person}{Ross
  Girshick}.} \bibinfo{year}{2018}\natexlab{}.
\newblock \bibinfo{title}{Mask R-CNN}.
\newblock
\newblock
\showeprint[arxiv]{1703.06870}~[cs.CV]


\bibitem[Hu et~al\mbox{.}(2023)]%
        {hu2023bliva}
\bibfield{author}{\bibinfo{person}{Wenbo Hu}, \bibinfo{person}{Yifan Xu},
  \bibinfo{person}{Yi Li}, \bibinfo{person}{Weiyue Li}, \bibinfo{person}{Zeyuan
  Chen}, {and} \bibinfo{person}{Zhuowen Tu}.} \bibinfo{year}{2023}\natexlab{}.
\newblock \bibinfo{title}{BLIVA: A Simple Multimodal LLM for Better Handling of
  Text-Rich Visual Questions}.
\newblock
\newblock
\showeprint[arxiv]{2308.09936}~[cs.CV]


\bibitem[Huang et~al\mbox{.}(2019)]%
        {icdar_19}
\bibfield{author}{\bibinfo{person}{Zheng Huang}, \bibinfo{person}{Kai Chen},
  \bibinfo{person}{Jianhua He}, \bibinfo{person}{Xiang Bai},
  \bibinfo{person}{Dimosthenis Karatzas}, \bibinfo{person}{Shijian Lu}, {and}
  \bibinfo{person}{C.~V. Jawahar}.} \bibinfo{year}{2019}\natexlab{}.
\newblock \showarticletitle{ICDAR2019 Competition on Scanned Receipt OCR and
  Information Extraction}. In \bibinfo{booktitle}{\emph{2019 International
  Conference on Document Analysis and Recognition (ICDAR)}}.
  \bibinfo{pages}{1516--1520}.
\newblock
\urldef\tempurl%
\url{https://doi.org/10.1109/ICDAR.2019.00244}
\showDOI{\tempurl}


\bibitem[Jaderberg et~al\mbox{.}(2016)]%
        {jaderberg_16}
\bibfield{author}{\bibinfo{person}{Max Jaderberg}, \bibinfo{person}{Karen
  Simonyan}, \bibinfo{person}{Andrea Vedaldi}, {and} \bibinfo{person}{Andrew
  Zisserman}.} \bibinfo{year}{2016}\natexlab{}.
\newblock \showarticletitle{Reading Text in the Wild with Convolutional Neural
  Networks}.
\newblock \bibinfo{journal}{\emph{Int. J. Comput. Vision}}
  \bibinfo{volume}{116}, \bibinfo{number}{1} (\bibinfo{date}{jan}
  \bibinfo{year}{2016}), \bibinfo{pages}{1–20}.
\newblock
\showISSN{0920-5691}
\urldef\tempurl%
\url{https://doi.org/10.1007/s11263-015-0823-z}
\showDOI{\tempurl}


\bibitem[Liao et~al\mbox{.}(2019)]%
        {mts2019}
\bibfield{author}{\bibinfo{person}{Minghui Liao}, \bibinfo{person}{Pengyuan
  Lyu}, \bibinfo{person}{Minghang He}, \bibinfo{person}{Cong Yao},
  \bibinfo{person}{Wenhao Wu}, {and} \bibinfo{person}{Xiang Bai}.}
  \bibinfo{year}{2019}\natexlab{}.
\newblock \bibinfo{title}{Mask TextSpotter: An End-to-End Trainable Neural
  Network for Spotting Text with Arbitrary Shapes}.
\newblock
\newblock
\showeprint[arxiv]{1908.08207}~[cs.CV]


\bibitem[Liu et~al\mbox{.}(2023b)]%
        {liu2023visual}
\bibfield{author}{\bibinfo{person}{Haotian Liu}, \bibinfo{person}{Chunyuan Li},
  \bibinfo{person}{Qingyang Wu}, {and} \bibinfo{person}{Yong~Jae Lee}.}
  \bibinfo{year}{2023}\natexlab{b}.
\newblock \bibinfo{title}{Visual Instruction Tuning}.
\newblock
\newblock
\showeprint[arxiv]{2304.08485}~[cs.CV]


\bibitem[Liu et~al\mbox{.}(2018)]%
        {pan2018}
\bibfield{author}{\bibinfo{person}{Shu Liu}, \bibinfo{person}{Lu Qi},
  \bibinfo{person}{Haifang Qin}, \bibinfo{person}{Jianping Shi}, {and}
  \bibinfo{person}{Jiaya Jia}.} \bibinfo{year}{2018}\natexlab{}.
\newblock \bibinfo{title}{Path Aggregation Network for Instance Segmentation}.
\newblock
\newblock
\showeprint[arxiv]{1803.01534}~[cs.CV]


\bibitem[Liu et~al\mbox{.}(2023a)]%
        {liu2023hidden}
\bibfield{author}{\bibinfo{person}{Yuliang Liu}, \bibinfo{person}{Zhang Li},
  \bibinfo{person}{Hongliang Li}, \bibinfo{person}{Wenwen Yu},
  \bibinfo{person}{Yang Liu}, \bibinfo{person}{Biao Yang},
  \bibinfo{person}{Mingxin Huang}, \bibinfo{person}{Dezhi Peng},
  \bibinfo{person}{Mingyu Liu}, \bibinfo{person}{Mingrui Chen},
  \bibinfo{person}{Chunyuan Li}, \bibinfo{person}{Xucheng Yin},
  \bibinfo{person}{Cheng lin Liu}, \bibinfo{person}{Lianwen Jin}, {and}
  \bibinfo{person}{Xiang Bai}.} \bibinfo{year}{2023}\natexlab{a}.
\newblock \bibinfo{title}{On the Hidden Mystery of OCR in Large Multimodal
  Models}.
\newblock
\newblock
\showeprint[arxiv]{2305.07895}~[cs.CV]


\bibitem[Long et~al\mbox{.}(2022a)]%
        {gocr2022}
\bibfield{author}{\bibinfo{person}{Shangbang Long}, \bibinfo{person}{Siyang
  Qin}, \bibinfo{person}{Dmitry Panteleev}, \bibinfo{person}{Alessandro
  Bissacco}, \bibinfo{person}{Yasuhisa Fujii}, {and} \bibinfo{person}{Michalis
  Raptis}.} \bibinfo{year}{2022}\natexlab{a}.
\newblock \bibinfo{title}{Towards End-to-End Unified Scene Text Detection and
  Layout Analysis}.
\newblock
\newblock
\showeprint[arxiv]{2203.15143}~[cs.CV]


\bibitem[Long et~al\mbox{.}(2022b)]%
        {heirtext}
\bibfield{author}{\bibinfo{person}{Shangbang Long}, \bibinfo{person}{Siyang
  Qin}, \bibinfo{person}{Dmitry Panteleev}, \bibinfo{person}{Alessandro
  Bissacco}, \bibinfo{person}{Yasuhisa Fujii}, {and} \bibinfo{person}{Michalis
  Raptis}.} \bibinfo{year}{2022}\natexlab{b}.
\newblock \showarticletitle{Towards End-to-End Unified Scene Text Detection and
  Layout Analysis}. In \bibinfo{booktitle}{\emph{2022 IEEE/CVF Conference on
  Computer Vision and Pattern Recognition (CVPR)}}.
  \bibinfo{pages}{1039--1049}.
\newblock
\urldef\tempurl%
\url{https://doi.org/10.1109/CVPR52688.2022.00112}
\showDOI{\tempurl}


\bibitem[Ma et~al\mbox{.}(2018)]%
        {rrpn2018}
\bibfield{author}{\bibinfo{person}{Jianqi Ma}, \bibinfo{person}{Weiyuan Shao},
  \bibinfo{person}{Hao Ye}, \bibinfo{person}{Li Wang}, \bibinfo{person}{Hong
  Wang}, \bibinfo{person}{Yingbin Zheng}, {and} \bibinfo{person}{Xiangyang
  Xue}.} \bibinfo{year}{2018}\natexlab{}.
\newblock \showarticletitle{Arbitrary-Oriented Scene Text Detection via
  Rotation Proposals}.
\newblock \bibinfo{journal}{\emph{IEEE Transactions on Multimedia}}
  \bibinfo{volume}{20}, \bibinfo{number}{11} (\bibinfo{date}{Nov.}
  \bibinfo{year}{2018}), \bibinfo{pages}{3111–3122}.
\newblock
\showISSN{1941-0077}
\urldef\tempurl%
\url{https://doi.org/10.1109/tmm.2018.2818020}
\showDOI{\tempurl}


\bibitem[Mathew et~al\mbox{.}(2021)]%
        {docvqa_21}
\bibfield{author}{\bibinfo{person}{Minesh Mathew}, \bibinfo{person}{Dimosthenis
  Karatzas}, {and} \bibinfo{person}{C.~V. Jawahar}.}
  \bibinfo{year}{2021}\natexlab{}.
\newblock \showarticletitle{DocVQA: A Dataset for VQA on Document Images}. In
  \bibinfo{booktitle}{\emph{2021 IEEE Winter Conference on Applications of
  Computer Vision (WACV)}}. \bibinfo{pages}{2199--2208}.
\newblock
\urldef\tempurl%
\url{https://doi.org/10.1109/WACV48630.2021.00225}
\showDOI{\tempurl}


\bibitem[Munjal et~al\mbox{.}(2021)]%
        {stride_21}
\bibfield{author}{\bibinfo{person}{Rachit~S Munjal}, \bibinfo{person}{Arun~D
  Prabhu}, \bibinfo{person}{Nikhil Arora}, \bibinfo{person}{Sukumar Moharana},
  {and} \bibinfo{person}{Gopi Ramena}.} \bibinfo{year}{2021}\natexlab{}.
\newblock \showarticletitle{STRIDE: Scene Text Recognition In-Device}. In
  \bibinfo{booktitle}{\emph{2021 International Joint Conference on Neural
  Networks (IJCNN)}}. \bibinfo{pages}{1--8}.
\newblock
\urldef\tempurl%
\url{https://doi.org/10.1109/IJCNN52387.2021.9534319}
\showDOI{\tempurl}


\bibitem[Nagel et~al\mbox{.}(2021)]%
        {white2021}
\bibfield{author}{\bibinfo{person}{Markus Nagel}, \bibinfo{person}{Marios
  Fournarakis}, \bibinfo{person}{Rana~Ali Amjad}, \bibinfo{person}{Yelysei
  Bondarenko}, \bibinfo{person}{Mart van Baalen}, {and} \bibinfo{person}{Tijmen
  Blankevoort}.} \bibinfo{year}{2021}\natexlab{}.
\newblock \bibinfo{title}{A White Paper on Neural Network Quantization}.
\newblock
\newblock
\showeprint[arxiv]{2106.08295}~[cs.LG]


\bibitem[Ren et~al\mbox{.}(2016)]%
        {faster2016}
\bibfield{author}{\bibinfo{person}{Shaoqing Ren}, \bibinfo{person}{Kaiming He},
  \bibinfo{person}{Ross Girshick}, {and} \bibinfo{person}{Jian Sun}.}
  \bibinfo{year}{2016}\natexlab{}.
\newblock \bibinfo{title}{Faster R-CNN: Towards Real-Time Object Detection with
  Region Proposal Networks}.
\newblock
\newblock
\showeprint[arxiv]{1506.01497}~[cs.CV]


\bibitem[Shenoy et~al\mbox{.}(2021a)]%
        {shenoy-etal-2021-asr}
\bibfield{author}{\bibinfo{person}{Ashish Shenoy}, \bibinfo{person}{Sravan
  Bodapati}, {and} \bibinfo{person}{Katrin Kirchhoff}.}
  \bibinfo{year}{2021}\natexlab{a}.
\newblock \showarticletitle{{ASR} Adaptation for {E}-commerce Chatbots using
  Cross-Utterance Context and Multi-Task Language Modeling}. In
  \bibinfo{booktitle}{\emph{Proceedings of the 4th Workshop on e-Commerce and
  NLP}}, \bibfield{editor}{\bibinfo{person}{Shervin Malmasi},
  \bibinfo{person}{Surya Kallumadi}, \bibinfo{person}{Nicola Ueffing},
  \bibinfo{person}{Oleg Rokhlenko}, \bibinfo{person}{Eugene Agichtein}, {and}
  \bibinfo{person}{Ido Guy}} (Eds.). \bibinfo{publisher}{Association for
  Computational Linguistics}, \bibinfo{address}{Online},
  \bibinfo{pages}{18--25}.
\newblock
\urldef\tempurl%
\url{https://doi.org/10.18653/v1/2021.ecnlp-1.3}
\showDOI{\tempurl}


\bibitem[Shenoy et~al\mbox{.}(2021b)]%
        {shenoy21_interspeech}
\bibfield{author}{\bibinfo{person}{Ashish Shenoy}, \bibinfo{person}{Sravan
  Bodapati}, \bibinfo{person}{Monica Sunkara}, \bibinfo{person}{Srikanth
  Ronanki}, {and} \bibinfo{person}{Katrin Kirchhoff}.}
  \bibinfo{year}{2021}\natexlab{b}.
\newblock \showarticletitle{{Adapting Long Context NLM for ASR Rescoring in
  Conversational Agents}}. In \bibinfo{booktitle}{\emph{Proc. Interspeech
  2021}}. \bibinfo{pages}{3246--3250}.
\newblock
\showISSN{2308-457X}
\urldef\tempurl%
\url{https://doi.org/10.21437/Interspeech.2021-1849}
\showDOI{\tempurl}


\bibitem[Shi et~al\mbox{.}(2017)]%
        {paddle_ocr_17}
\bibfield{author}{\bibinfo{person}{Baoguang Shi}, \bibinfo{person}{Xiang Bai},
  {and} \bibinfo{person}{Cong Yao}.} \bibinfo{year}{2017}\natexlab{}.
\newblock \showarticletitle{An End-to-End Trainable Neural Network for
  Image-Based Sequence Recognition and Its Application to Scene Text
  Recognition}.
\newblock \bibinfo{journal}{\emph{IEEE Trans. Pattern Anal. Mach. Intell.}}
  \bibinfo{volume}{39}, \bibinfo{number}{11} (\bibinfo{date}{nov}
  \bibinfo{year}{2017}), \bibinfo{pages}{2298–2304}.
\newblock
\showISSN{0162-8828}
\urldef\tempurl%
\url{https://doi.org/10.1109/TPAMI.2016.2646371}
\showDOI{\tempurl}


\bibitem[Shi et~al\mbox{.}(2016)]%
        {shi_robust_str_16}
\bibfield{author}{\bibinfo{person}{Baoguang Shi}, \bibinfo{person}{Xinggang
  Wang}, \bibinfo{person}{Pengyuan Lyu}, \bibinfo{person}{Cong Yao}, {and}
  \bibinfo{person}{Xiang Bai}.} \bibinfo{year}{2016}\natexlab{}.
\newblock \showarticletitle{Robust Scene Text Recognition with Automatic
  Rectification}. In \bibinfo{booktitle}{\emph{2016 IEEE Conference on Computer
  Vision and Pattern Recognition (CVPR)}}. \bibinfo{pages}{4168--4176}.
\newblock
\urldef\tempurl%
\url{https://doi.org/10.1109/CVPR.2016.452}
\showDOI{\tempurl}


\bibitem[Shi et~al\mbox{.}(2023)]%
        {shi2023exploring}
\bibfield{author}{\bibinfo{person}{Yongxin Shi}, \bibinfo{person}{Dezhi Peng},
  \bibinfo{person}{Wenhui Liao}, \bibinfo{person}{Zening Lin},
  \bibinfo{person}{Xinhong Chen}, \bibinfo{person}{Chongyu Liu},
  \bibinfo{person}{Yuyi Zhang}, {and} \bibinfo{person}{Lianwen Jin}.}
  \bibinfo{year}{2023}\natexlab{}.
\newblock \bibinfo{title}{Exploring OCR Capabilities of GPT-4V(ision) : A
  Quantitative and In-depth Evaluation}.
\newblock
\newblock
\showeprint[arxiv]{2310.16809}~[cs.CV]


\bibitem[Singh et~al\mbox{.}(2021)]%
        {textocr2021}
\bibfield{author}{\bibinfo{person}{Amanpreet Singh}, \bibinfo{person}{Guan
  Pang}, \bibinfo{person}{Mandy Toh}, \bibinfo{person}{Jing Huang},
  \bibinfo{person}{Wojciech Galuba}, {and} \bibinfo{person}{Tal Hassner}.}
  \bibinfo{year}{2021}\natexlab{}.
\newblock \bibinfo{title}{TextOCR: Towards large-scale end-to-end reasoning for
  arbitrary-shaped scene text}.
\newblock
\newblock
\showeprint[arxiv]{2105.05486}~[cs.CV]


\bibitem[Team et~al\mbox{.}(2023)]%
        {geminiteam2023gemini}
\bibfield{author}{\bibinfo{person}{Gemini Team}, \bibinfo{person}{Rohan Anil},
  \bibinfo{person}{Sebastian Borgeaud}, \bibinfo{person}{Yonghui Wu},
  \bibinfo{person}{Jean-Baptiste Alayrac}, \bibinfo{person}{Jiahui Yu},
  \bibinfo{person}{Radu Soricut}, \bibinfo{person}{Johan Schalkwyk},
  \bibinfo{person}{Andrew~M. Dai}, {and} \bibinfo{person}{et al}.}
  \bibinfo{year}{2023}\natexlab{}.
\newblock \bibinfo{title}{Gemini: A Family of Highly Capable Multimodal
  Models}.
\newblock
\newblock
\showeprint[arxiv]{2312.11805}~[cs.CL]


\bibitem[Tian et~al\mbox{.}(2019)]%
        {fcos2019}
\bibfield{author}{\bibinfo{person}{Zhi Tian}, \bibinfo{person}{Chunhua Shen},
  \bibinfo{person}{Hao Chen}, {and} \bibinfo{person}{Tong He}.}
  \bibinfo{year}{2019}\natexlab{}.
\newblock \bibinfo{title}{FCOS: Fully Convolutional One-Stage Object
  Detection}.
\newblock
\newblock
\showeprint[arxiv]{1904.01355}~[cs.CV]


\bibitem[Wan et~al\mbox{.}(2020)]%
        {Wan2020FBNetV2DN}
\bibfield{author}{\bibinfo{person}{Alvin Wan}, \bibinfo{person}{Xiaoliang Dai},
  \bibinfo{person}{Peizhao Zhang}, \bibinfo{person}{Zijian He},
  \bibinfo{person}{Yuandong Tian}, \bibinfo{person}{Saining Xie},
  \bibinfo{person}{Bichen Wu}, \bibinfo{person}{Matthew Yu},
  \bibinfo{person}{Tao Xu}, \bibinfo{person}{Kan Chen},
  \bibinfo{person}{P{\'e}ter Vajda}, {and} \bibinfo{person}{Joseph Gonzalez}.}
  \bibinfo{year}{2020}\natexlab{}.
\newblock \showarticletitle{FBNetV2: Differentiable Neural Architecture Search
  for Spatial and Channel Dimensions}.
\newblock \bibinfo{journal}{\emph{2020 IEEE/CVF Conference on Computer Vision
  and Pattern Recognition (CVPR)}} (\bibinfo{year}{2020}),
  \bibinfo{pages}{12962--12971}.
\newblock
\urldef\tempurl%
\url{https://api.semanticscholar.org/CorpusID:215744832}
\showURL{%
\tempurl}


\bibitem[Wang et~al\mbox{.}(2021)]%
        {maxdeeplab21}
\bibfield{author}{\bibinfo{person}{Huiyu Wang}, \bibinfo{person}{Yukun Zhu},
  \bibinfo{person}{Hartwig Adam}, \bibinfo{person}{Alan Yuille}, {and}
  \bibinfo{person}{Liang-Chieh Chen}.} \bibinfo{year}{2021}\natexlab{}.
\newblock \bibinfo{title}{MaX-DeepLab: End-to-End Panoptic Segmentation with
  Mask Transformers}.
\newblock
\newblock
\showeprint[arxiv]{2012.00759}~[cs.CV]


\bibitem[Wang and Belongie(2010)]%
        {wang_belongie_10}
\bibfield{author}{\bibinfo{person}{Kai Wang} {and} \bibinfo{person}{Serge
  Belongie}.} \bibinfo{year}{2010}\natexlab{}.
\newblock \showarticletitle{Word Spotting in the Wild}. In
  \bibinfo{booktitle}{\emph{Proceedings of the 11th European Conference on
  Computer Vision: Part I}} (Heraklion, Crete, Greece)
  \emph{(\bibinfo{series}{ECCV'10})}. \bibinfo{publisher}{Springer-Verlag},
  \bibinfo{address}{Berlin, Heidelberg}, \bibinfo{pages}{591–604}.
\newblock
\showISBNx{3642155480}


\bibitem[Yang et~al\mbox{.}(2022)]%
        {kld2022}
\bibfield{author}{\bibinfo{person}{Xue Yang}, \bibinfo{person}{Xiaojiang Yang},
  \bibinfo{person}{Jirui Yang}, \bibinfo{person}{Qi Ming},
  \bibinfo{person}{Wentao Wang}, \bibinfo{person}{Qi Tian}, {and}
  \bibinfo{person}{Junchi Yan}.} \bibinfo{year}{2022}\natexlab{}.
\newblock \bibinfo{title}{Learning High-Precision Bounding Box for Rotated
  Object Detection via Kullback-Leibler Divergence}.
\newblock
\newblock
\showeprint[arxiv]{2106.01883}~[cs.CV]


\bibitem[Ye et~al\mbox{.}(2023)]%
        {ye2023ureader}
\bibfield{author}{\bibinfo{person}{Jiabo Ye}, \bibinfo{person}{Anwen Hu},
  \bibinfo{person}{Haiyang Xu}, \bibinfo{person}{Qinghao Ye},
  \bibinfo{person}{Ming Yan}, \bibinfo{person}{Guohai Xu},
  \bibinfo{person}{Chenliang Li}, \bibinfo{person}{Junfeng Tian},
  \bibinfo{person}{Qi Qian}, \bibinfo{person}{Ji Zhang}, \bibinfo{person}{Qin
  Jin}, \bibinfo{person}{Liang He}, \bibinfo{person}{Xin~Alex Lin}, {and}
  \bibinfo{person}{Fei Huang}.} \bibinfo{year}{2023}\natexlab{}.
\newblock \bibinfo{title}{UReader: Universal OCR-free Visually-situated
  Language Understanding with Multimodal Large Language Model}.
\newblock
\newblock
\showeprint[arxiv]{2310.05126}~[cs.CV]


\bibitem[Zhang et~al\mbox{.}(2021)]%
        {varifocal2021}
\bibfield{author}{\bibinfo{person}{Haoyang Zhang}, \bibinfo{person}{Ying Wang},
  \bibinfo{person}{Feras Dayoub}, {and} \bibinfo{person}{Niko Sünderhauf}.}
  \bibinfo{year}{2021}\natexlab{}.
\newblock \bibinfo{title}{VarifocalNet: An IoU-aware Dense Object Detector}.
\newblock
\newblock
\showeprint[arxiv]{2008.13367}~[cs.CV]


\bibitem[Zhu et~al\mbox{.}(2023)]%
        {zhu2023minigpt4}
\bibfield{author}{\bibinfo{person}{Deyao Zhu}, \bibinfo{person}{Jun Chen},
  \bibinfo{person}{Xiaoqian Shen}, \bibinfo{person}{Xiang Li}, {and}
  \bibinfo{person}{Mohamed Elhoseiny}.} \bibinfo{year}{2023}\natexlab{}.
\newblock \bibinfo{title}{MiniGPT-4: Enhancing Vision-Language Understanding
  with Advanced Large Language Models}.
\newblock
\newblock
\showeprint[arxiv]{2304.10592}~[cs.CV]


\end{thebibliography}

\appendix

\pagebreak
\section{Appendix}

\begin{figure}[H]
    \centering
    \includegraphics[width=0.3\textwidth]{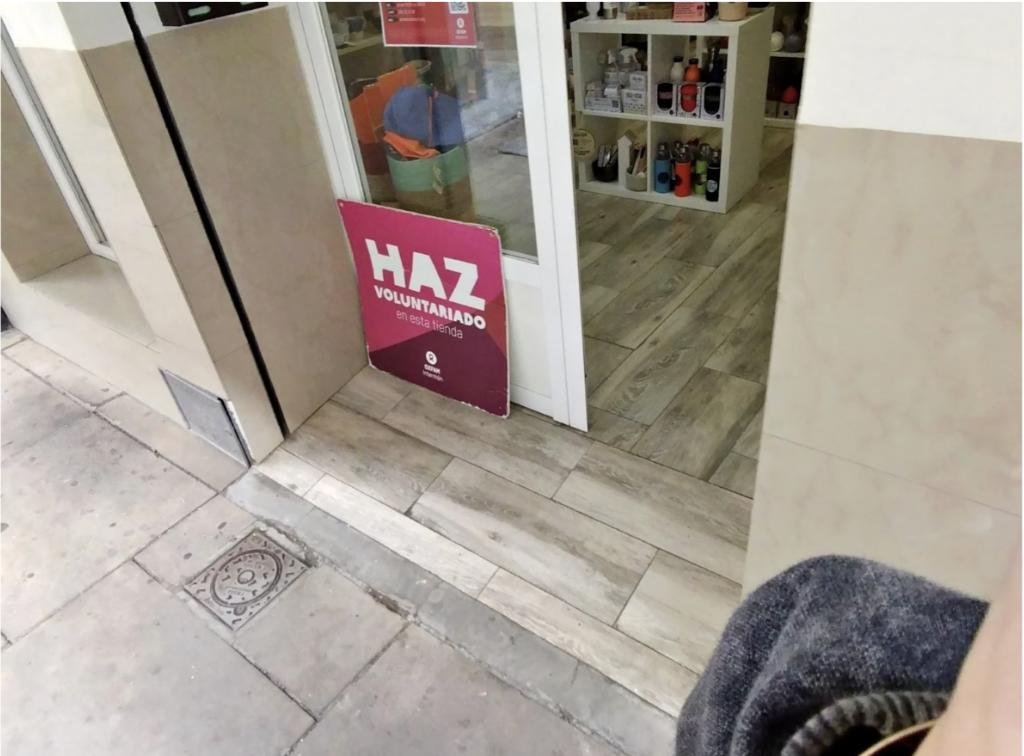}
    \caption{Sample input image for \textsc{Lumos}}
    \label{fig:prompt_example}
\end{figure}

\begin{figure}[H]
    \centering
    \includegraphics[width=0.3\textwidth]{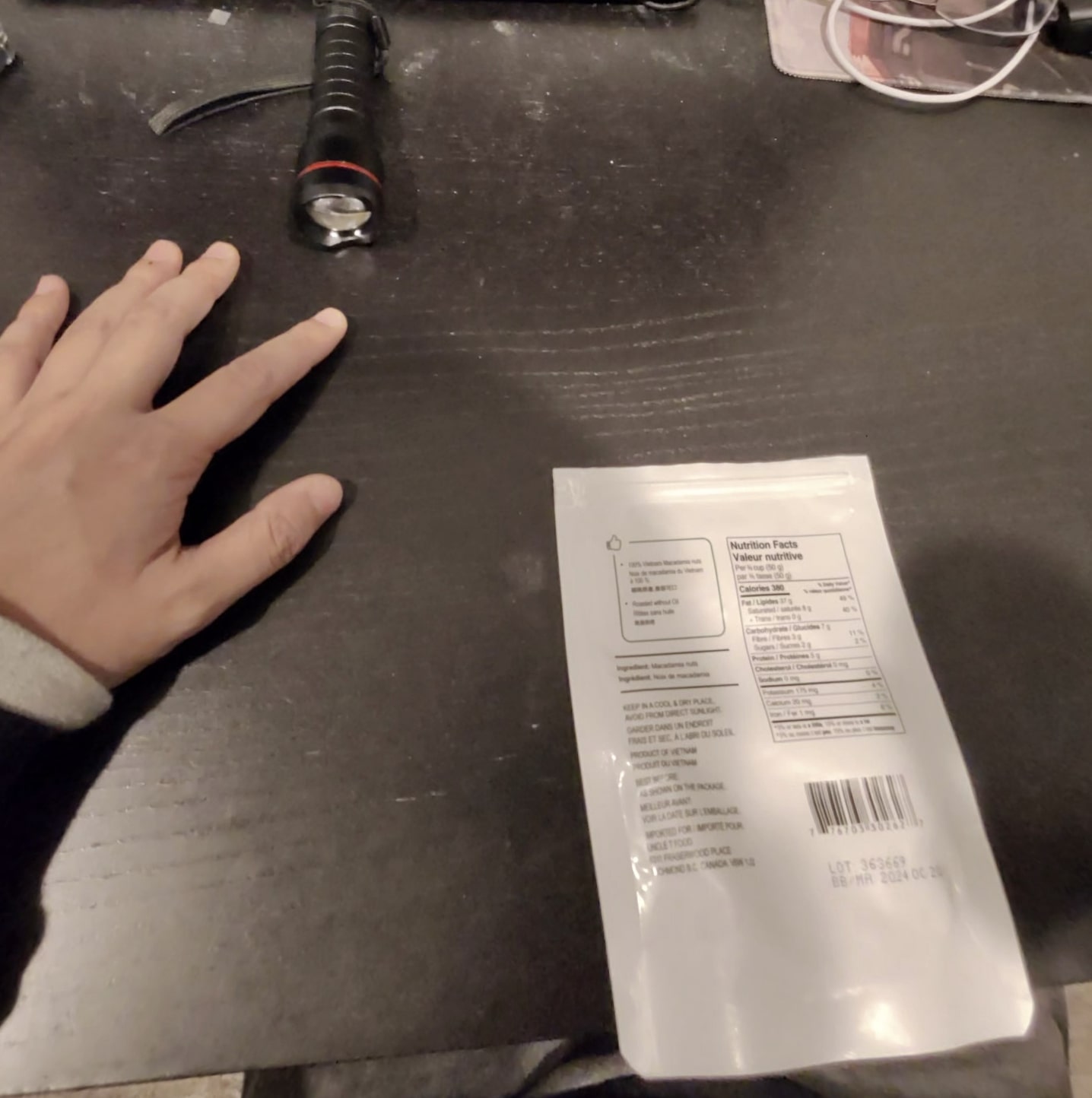}
    \caption{Example of a hand that is not indicative of ROI}
    \label{fig:fp_hand}
\end{figure}
\begin{figure}[H]
    \includegraphics[width=0.35\textwidth]{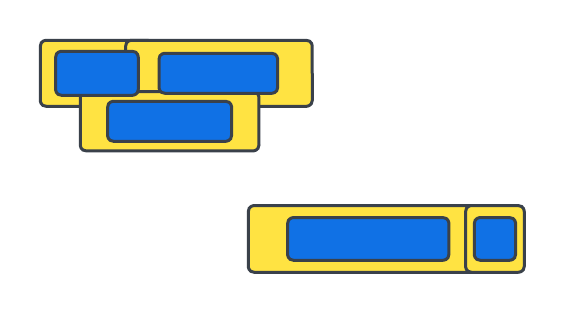}
    \caption{How we construct paragraphs: Blue boxes are from text detection which is then expanded (yellow). There are two paragraphs based on how the yellow boxes overlap}
    \label{expandbox}
\end{figure}

\begin{algorithm*}
    \caption{Reading Order Reconstruction Algorithm}
  \begin{algorithmic}[1]
\STATE \textbf{Input} Recognized word $w_1,w_2...w_n$,their corresponding bounding box $b_1,b_2...b_n$, vertical expansion ratio $r_v$, horizontal expansion ratio $r_h$,iou threshold $t$
\STATE \textbf{Step 1} Expand all boxes $b_1,b_2...b_n$ by $r_v$ vertically and $r_h$ horizontally to get expanded boxes $\hat{b}_1,\hat{b}_2...\hat{b}_n$ as shown in figure \ref{expandbox}
\STATE \textbf{Step 2} Compute pairwise IOU of  $\hat{b}_1,\hat{b}_2...\hat{b}_n$, construct adjacency matrix $A\in [0,1]^{n\times n}$ where each element of $A$ is defined as 
\[
    a_{i,j}= 
\begin{cases}
    1,& \text{if } iou(\hat{b}_i,\hat{b}_j)\geq t\\
    0,              & \text{otherwise}
\end{cases}
\]
\STATE \textbf{Step 3} Finding connected components 
\[C_1=\{1_1,1_2,...\},...,C_m=\{m_1,m_2,...\}\] 
from the graph defined by the adjacency matrix $A$ (each word box is a vertex) where $k_l$ is the index of the $l^{th}$ box in the $k^{th}$ connected component. Each connect component stands for a paragraph.
\STATE \textbf{Step 4} Applying raster scan to words within each paragraph (connected components) to generate words in reading order for each paragraph.
\STATE \textbf{Step 5} Compute the location of the paragraph by finding the minimum area rectangle enclosing all words in each connects components
\STATE \textbf{Output} Words in each paragraphs in reading order: $P_1=\{w_{1_1},w_{1_2},..\}, ..., P_m=\{w_{m_1},w_{m_2}\}$, \\
Locations of the rectangular area for each paragraph in the format of $(top,left,height,width)$.
  \end{algorithmic}
  \label{grouping_alg}
\end{algorithm*}

\begin{table*}[ht]
\centering
\caption{Sample prompt to different variants of \textsc{Lumos} for Figure~\ref{fig:prompt_example}}
\label{tab:prompt}
\begin{tabular}{lllll}
\toprule
&&&\\
System & Prompt & Response & Relevant \\
\midrule
MM-LLM only&What does this sign say & The sign says, "Haz Matado" &irrelevant\\
&&which translates to "You've killed it." &   \\
\midrule
MM-LLM+STR &Here is a transcription of the recognized &The sign says, "HAZ&\\
&text in the image, which may contain&VOLUNTARIADO en esta tienda" which & relevant \\
& misspelled words. [HAZ VOLUNTARIADO &translates to "DO VOLUNTEER WORK&\\
&en esta tienda R OKFAM intermon]. &this store"&\\
&My question is what&&\\
&does this sign sayin &&\\
\midrule
MM-LLM+STR+Position & I have included the coordinates of the &The sign says, "HAZ&\\
&text within the image. The coordinates&VOLUNTARIADO en esta tienda" which & relevant \\
&(0,0) is indicative of the top-left &translates to "DO VOLUNTEER WORK&\\
&corner of the image and (3024, 4032) &in this store"&\\
&is indicative of the bottom right corner&&\\
&of the image. For example, the tuple&&\\
&(zzzz, 10, 20) means the paragraph&&\\
&zzzz is centered at the point (10, 20)&&\\
&within the image. Here is a transcription&&\\
&of the recognized text in the image&&\\
&according to this coordinate system.&&\\
&It may contain misspelled words.&&\\
&[(HAZ VOLUNTARIADO en esta tienda&&\\
& , 1260, 1532) (R OKFAM intermon,&&\\
&1282, 1939)]. My question is what &&\\
&does this sign say. &&\\
\bottomrule
\end{tabular}
\vspace{-1.0em}
\end{table*}

\end{document}